\documentclass[11pt]{article}

\usepackage[english]{babel}
\usepackage[utf8]{inputenc}
\usepackage{fullpage}

\usepackage[numbers]{natbib}
\usepackage{amsmath}
\usepackage{amssymb}
\usepackage{amsthm}
    \newtheorem{definition}{Definition}
    \newtheorem{theorem}{Theorem}
\usepackage{mathtools} 
    \DeclarePairedDelimiter\abs{\lvert}{\rvert}%
\usepackage[boxruled]{algorithm2e}
    \newcommand{\balg}{\begin{algorithm}[htbp]\DontPrintSemicolon}
    \newcommand{\ealg}{\end{algorithm}}

\usepackage{booktabs}
\usepackage{caption, subcaption}
\usepackage{graphicx}
\usepackage{hyperref}

\newcommand{\imgwidth}{.85\textwidth}
\newlength{\tablewidth}
\title{%
    A novel initialisation based on hospital-resident assignment for the
    \(k\)-modes algorithm
}
\author{Henry Wilde, Vincent Knight and Jonathan Gillard}

\begin{document}

\maketitle%
\begin{abstract}
    This paper presents a new way of selecting an initial solution for the
    \(k\)-modes algorithm that allows for a notion of mathematical fairness and
    a leverage of the data that the common initialisations from literature do
    not. The method, which utilises the Hospital-Resident Assignment Problem to
    find the set of initial cluster centroids, is compared with the current
    initialisations on both benchmark datasets and a body of newly generated
    artificial datasets. Based on this analysis, the proposed method is shown to
    outperform the other initialisations in the majority of cases, especially
    when the number of clusters is optimised. In addition, we find that our
    method outperforms the leading established method specifically for
    low-density data.
\end{abstract}

\section{Introduction}\label{sec:intro}

This work focusses on \(k\)-modes clustering --- an extension to
\(k\)-means that permits the sensible clustering of categorical (i.e.\ ordinal,
nominal or otherwise discrete) data as set out in the seminal works by
Huang~\cite{Huang1997a,Huang1997b,Huang1998}. In particular, the interest of
this paper is in how the performance of the \(k\)-modes algorithm is affected by
the quality of its initial solution. The initialisation method proposed in this
work extends the method presented by Huang~\cite{Huang1998} by using results
from game theory to ensure mathematical fairness and to lever the full learning
opportunities presented by the data being clustered. In doing so, it is
demonstrated that the proposed method is able to outperform both of the
established initialisations for \(k\)-modes. The paper is structured as follows:
\begin{itemize}
    \item Section~\ref{sec:intro} introduces the \(k\)-modes algorithm and its
        established initialisation methods.
    \item Section~\ref{sec:method} provides a brief overview of
        matching games and their variants before a statement of the proposed
        initialisation.
    \item Section~\ref{sec:results} presents analyses of the initialisations
        on benchmark and new, artificial datasets.
    \item Section~\ref{sec:conclusion} concludes the paper.
\end{itemize}

\subsection{The \(k\)-modes algorithm}\label{subsec:kmodes}

The following notation will be used throughout this work to describe the objects
associated with clustering a categorical dataset:

\begin{itemize}
    \item Let \(\mathcal{A} := A_1 \times \cdots \times A_m\) denote the
        \emph{attribute~space}. In this work, only categorical attributes are
        considered, i.e.\ for each \(j = 1, \ldots, m\) it follows that \(A_j :=
        \left\{a_1^{(j)}, \ldots, a_{d_j}^{(j)}\right\}\) where \(d_j = |A_j|\)
        is the size of the \(j^{th}\) attribute.

    \item Let \(\mathcal{X} := \left\{X^{(1)}, \ldots, X^{(N)}\right\} \subset
        \mathcal{A}\) denote a \emph{dataset} where each \(X^{(i)} \in
        \mathcal{X}\) is defined as an \(m\)-tuple \(X^{(i)} := \left(x_1^{(i)},
        \ldots, x_m^{(i)}\right)\) where \(x_j^{(i)} \in A_j\) for each \(j = 1,
        \ldots, m\). The elements of \(\mathcal{X}\) are referred to as
        \emph{data points} or \emph{instances}.
    \item Let \(\mathcal{Z} := \left(Z_1, \ldots, Z_k\right)\) be a partition
        of a dataset \(\mathcal{X} \subset \mathcal A\) into \(k \in
        \mathbb{Z}^{+}\) distinct, non-empty parts. Such a partition
        \(\mathcal{Z}\) is called a \emph{clustering} of \(\mathcal{X}\).

    \item Each cluster \(Z_l\) has associated with it a
        \emph{mode} (see Definition~\ref{def:mode}) which is
        denoted by \(z^{(l)} = \left(z_1^{(l)},~\ldots,~z_m^{(l)}\right) \in
        \mathcal{A}\).  These points are also referred to as
        \emph{representative~points} or \emph{centroids}. The set of all current
        cluster modes is denoted as \(\overline Z = \left\{z^{(1)}, \ldots,
        z^{(k)}\right\}\).
\end{itemize}

Definition~\ref{def:dissim} describes a dissimilarity measure between
categorical data points.

\begin{definition}\label{def:dissim}
    Let \(\mathcal{X} \subset \mathcal A\) be a dataset and consider any
    \(X^{(a)}, X^{(b)} \in \mathcal{X}\). The dissimilarity between \(X^{(a)}\)
    and \(X^{(b)}\), denoted by \(d\left(X^{(a)}, X^{(b)}\right)\), is given by:
    \begin{equation}\label{eq:dissim}
        d\left(X^{(a)}, X^{(b)}\right) := \sum_{j=1}^{m} \delta\left(x_j^{(a)},
        x_j^{(b)}\right) \quad \text{where} \quad \delta\left(x, y\right) =
        \begin{cases}
            0, & \text{if} \ x = y \\
            1, & \text{otherwise.}
        \end{cases}
    \end{equation}
\end{definition}

With this metric, the notion of a representative point of a cluster is
addressed. With numeric data and \(k\)-means, such a point is taken to be the
mean of the points within the cluster. With categorical data, however, the mode
is used as the measure for central tendency. This follows from the concept of
dissimilarity in that the point that best represents (i.e.\ is closest to) those
in a cluster is one with the most frequent attribute values of the points in the
cluster. The following definitions and theorem formalise this and a method to
find such a point.

\begin{definition}\label{def:mode}
    Let \(\mathcal{X} \subset \mathcal{A}\) be a dataset and consider some point
    \(z = \left(z_1, \ldots, z_m\right) \in \mathcal{A}\). Then \(z\) is called
    a \emph{mode} of \(\mathcal{X}\) if it minimises the following:
    \begin{equation}\label{eq:summed-dissim}
        D\left(\mathcal{X}, z\right) = \sum_{i=1}^{N} d\left(X^{(i)}, z\right)
    \end{equation}
\end{definition}

\begin{definition}\label{def:rel-freq}
    Let \(\mathcal{X} \subset \mathcal{A}\) be a dataset. Then
    \(n\left(a_s^{(j)}\right)\) denotes the \emph{frequency} of the \(s^{th}\)
    category \(a_s^{(j)}\) of \(A_j\) in \(\mathcal{X}\), i.e.\ for each \(A_j
    \in \mathcal{A}\) and each \(s = 1, \ldots, d_j\):
    \begin{equation}
        n\left(a_s^{(j)}\right) := \abs*{%
            {\left\{X^{(i)} \in \mathcal{X}: x_j^{(i)} = a_s^{(j)}\right\}}
        }
    \end{equation}
	
    Furthermore, \(\frac{n\left(a_s^{(j)}\right)}{N}\) is called the
    \emph{relative~frequency} of category \(a_s^{(j)}\) in \(\mathcal{X}\).
\end{definition}

\begin{theorem}\label{thm:mode}
    Consider a dataset \(\mathcal{X} \subset \mathcal{A}\) and some \(U = (u_1,
    \ldots, u_m) \in \mathcal{A}\). Then \(D(\mathcal{X}, U)\) is minimised if
    and only if \(n\left(u_j\right) \geq n\left(a_s^{(j)}\right)\) for all
    \(s=1, \ldots, d_j\) for each \(j = 1, \ldots, m\).

    A proof of this theorem can be found in the Appendix of~\cite{Huang1998}.
\end{theorem}

Theorem~\ref{thm:mode} defines the process by which cluster modes are updated
in \(k\)-modes (see Algorithm~\ref{alg:update}), and so the final component from
the \(k\)-means paradigm to be configured is the objective (cost) function. This
function is defined in Definition~\ref{def:cost}, and following that a practical
statement of the \(k\)-modes algorithm is given in Algorithm~\ref{alg:kmodes} as
set out in~\cite{Huang1998}.

\begin{definition}\label{def:cost}
    Let \(\mathcal{Z} = \left\{Z_1, \ldots, Z_k\right\}\) be a clustering of a
    dataset \(\mathcal{X}\), and let \(\overline Z = \left\{z^{(1)},
    \ldots, z^{(k)}\right\}\) be the corresponding cluster modes. Then \(W =
    \left(w_{i, l}\right)\) is an \(N \times k\) \emph{partition~matrix} of
    \(\mathcal{X}\) such that:
    \[
        w_{i, l} = \begin{cases}
                     1, & \text{if} \ X^{(i)} \in Z_l\\
                     0, & \text{otherwise.}
                   \end{cases}
    \]

    With this, the \emph{cost~function} is defined to be the summed
    within-cluster dissimilarity:
    \begin{equation}\label{eq:cost}
        C\left(W, \overline Z\right) := \sum_{l=1}^{k} \sum_{i=1}^{N}
        \sum_{j=1}^{m} w_{i,l} \ \delta\left(x_j^{(i)}, z_j^{(l)}\right)
    \end{equation}
\end{definition}

\balg%
    \caption{The \(k\)-modes algorithm}\label{alg:kmodes}
    \KwIn{a dataset \(\mathcal{X}\), a number of clusters to form \(k\)}
    \KwOut{a clustering \(\mathcal{Z}\) of \(\mathcal{X}\)}

    Select \(k\) initial modes \(z^{(1)}, \ldots, z^{(k)} \in \mathcal{X}\)\;
    \(\overline Z \gets \left\{z^{(1)}, \ldots, z^{(k)}\right\}\)\;
    \(\mathcal{Z} \gets \left(\left\{z^{(1)}\right\}, \ldots,
    \left\{z^{(k)}\right\}\right)\)\;

    \For{\(X^{(i)} \in \mathcal{X}\)}{%
        \(Z_{l^*} \gets \textsc{SelectClosest}\left(X^{(i)}\right)\)\;
        \(Z_{l^*} \gets Z_{l^*} \cup \left\{X^{(i)}\right\}\)\;
        \(\textsc{Update}\left(z^{(l^*)}\right)\)\;
    }

    \Repeat{No point changes cluster}{%
        \For{\(X^{(i)} \in \textbf{X}\)}{%
            Let \(Z_l\) be the cluster \(X^{(i)}\) currently belongs to\;
            \(Z_{l^*} \gets \textsc{SelectClosest}\left(X^{(i)}\right)\)\;
            \If{\(l \neq l^*\)}{%
                \(Z_{l} \gets Z_{l} \setminus \left\{X^{(i)}\right\}\) and
                \(Z_{l^*} \gets Z_{l^*} \cup \left\{X^{(i)}\right\}\)\;
                \(\textsc{Update}\left(z^{(l)}\right)\) and
                \(\textsc{Update}\left(z^{(l^*)}\right)\)\;
            }
        }
    }
\ealg%

\balg%
    \caption{\textsc{SelectClosest}}\label{alg:select_closest}
    \KwIn{%
        a data point \(X^{(i)}\), a set of current clusters \(\mathcal{Z}\) and
        their modes \(\overline Z\)
    }
    \KwOut{the cluster whose mode is closest to the data point \(Z_{l^*}\)}

    Select \(z^{l^*} \in \overline Z\) that minimises:
    \(d\left(X^{(i)}, z_{l^*}\right)\)\;
    Find their associated cluster \(Z_{l^*}\)
\ealg%

\balg%
    \caption{\textsc{Update}}\label{alg:update}
    \KwIn{an attribute space \(\mathcal{A}\), a mode to update \(z^{(l)}\) and
    its cluster \(Z_l\)}
    \KwOut{an updated mode}

    Find \(z \in \mathcal{A}\) that minimises \(D(Z_l, z)\)\;
    \(z^{(l)} \gets z\)\;
\ealg%

\subsection{Initialisation processes}\label{subsec:inits}

The standard selection method to initialise \(k\)-modes is to randomly sample
\(k\) distinct points in the dataset. In all cases, the initial modes must be
points in the dataset to ensure that there are no empty clusters in the first
iteration of the algorithm. The remainder of this section describes two
well-established initialisation methods that aim to preemptively lever the
structure of the data at hand.

\subsubsection{Huang's method}\label{subsec:huang}

Amongst the original works by Huang, an alternative initialisation method was
presented that selects modes by distributing frequently occurring values from
the attribute space among \(k\) potential modes~\cite{Huang1998}. The process,
denoted as Huang's method, is described in full in Algorithm~\ref{alg:huang}.
Huang's method considers a set of potential modes, \(\widehat Z \subset \mathcal
A\), that is then replaced by the actual set of initial modes, \(\overline Z
\subset \mathcal X\). The statement of how the set of potential modes are formed
is ambiguous in the original paper --- as is alluded to in~\cite{Jiang2016}.
Here, as is done in practical implementations of \(k\)-modes, this has been
interpreted as being done via a weighted random sample (see
Algorithm~\ref{alg:potential_modes}).

\balg%
    \caption{Huang's method}\label{alg:huang}
    \KwIn{a dataset \(\mathcal{X} \subset \mathcal{A}\), a number of modes to
    find \(k\)}
    \KwOut{a set of \(k\) initial modes \(\overline Z\)}

    \(\overline Z \gets \emptyset\)\;
    \(\widehat Z \gets \textsc{SamplePotentialModes}\left(\mathcal{X}\right)\)\;

    \For{\(\hat{z} \in \widehat Z\)}{%
        Select \(X^{(i^*)} \in \mathcal{X} \setminus \overline Z\) that
        minimises \(d\left(X^{(i)}, \hat{z}\right)\)\;

        \(\overline Z \gets \overline Z \cup \left\{X^{(i^*)}\right\}\)
    }
\ealg%

\balg%
    \caption{\textsc{SamplePotentialModes}}\label{alg:potential_modes}
    \KwIn{%
        a dataset \(\mathcal{X} \subset \mathcal{A}\), a number of modes to find
        \(k\)
    }
    \KwOut{a set of \(k\) potential modes \(\widehat Z\)}

    \(\widehat Z \gets \emptyset\)\;
    \For{\(j = 1, \ldots, m\)}{%
        \For{\(s = 1, \ldots, d_j\)}{%
            Calculate \(\frac{n\left(a_s^{(j)}\right)}{N}\)
        }
    }

    \While{\(\abs*{\widehat Z} < k\)}{%
        Create an empty \(m\)-tuple \(\hat{z}^{(l)}\)\;
        \For{\(j = 1, \ldots, m\)}{%
            Sample \(a_{s^*}^{(j)}\) from \(A_j\) with respect to the
            relative frequencies of \(A_j\)\;
            \(\hat{z}_j^{(l)} \gets a_{s^*}^{(j)}\)
        }
        \(\widehat Z \gets \widehat Z \cup \left\{\hat{z}^{(l)}\right\}\)
    }
\ealg%

\subsubsection{Cao's method}\label{subsec:cao}

The second initialisation process that is widely used with \(k\)-modes is known
as Cao's method~\cite{Cao2009}. This method selects the initial modes according
to their density in the dataset whilst forcing dissimilarity between them.
Definition~\ref{def:density} formalises the concept of density and its
relationship to relative frequency. The method, which is described in
Algorithm~\ref{alg:cao}, is deterministic --- unlike Huang's method which relies
on random sampling.

\begin{definition}\label{def:density}	
    Consider a dataset
    \(\mathcal{X} \subset \mathcal{A} = \{A_1, \ldots, A_m\}\). Then the
    \emph{average~density} of any point \(X_i \in \mathcal{X}\) with respect to
    \(\mathcal{A}\) is defined~\cite{Cao2009} as:
    \begin{equation}\label{eq:density}
        \text{Dens}\left(X^{(i)}\right) = \frac{%
            \sum_{j=1}^m \text{Dens}_{j}\left(X^{(i)}\right)
        }{m}
        \quad \text{where} \quad
        \text{Dens}_{j}\left(X^{(i)}\right) = \frac{%
            \abs*{%
                \left\{X^{(t)} \in \mathcal{X} : x_j^{(i)} = x_j^{(t)}\right\}
            }
        }{N}
    \end{equation}

    Observe that:
    \[
        \abs*{\left\{X^{(t)} \in \mathcal{X} : x_j^{(i)} = x_j^{(t)}\right\}}%
        = n\left(x_j^{(i)}\right)%
        = \sum_{t=1}^N \left(1 - \delta\left(x_j^{(i)}, x_j^{(t)}\right)\right)
    \]

    And so, an alternative definition for~\eqref{eq:density} can be derived:
    \begin{equation}\label{eq:density-alt}
        \text{Dens}\left(X^{(i)}\right)
        = \frac{1}{mN} \sum_{j=1}^m \sum_{t=1}^N \left(%
            1 - \delta\left(x_j^{(i)}, x_j^{(t)}\right)
        \right)
        = 1 - \frac{1}{mN} D\left(\mathcal{X}, X^{(i)}\right)
    \end{equation}
\end{definition}

\balg%
    \caption{Cao's method}\label{alg:cao}
    \KwIn{a dataset \(\mathcal{X}\), a number of modes to find \(k\)}
    \KwOut{a set of \(k\) initial modes \(\overline Z\)}

    \(\overline Z \gets \emptyset\)\;
    \For{\(X^{(i)} \in \mathcal{X}\)}{%
        Calculate \(\text{Dens}\left(X^{(i)}\right)\)
    }

    Select \(1 \leq i_1 \leq N\) which maximises
    \(\text{Dens}\left(X^{(i)}\right)\)\;

    \(\overline Z \gets \overline Z \cup \left\{X^{(i_1)}\right\}\)\;

    \While{\(\abs*{\overline Z} < k\)}{%
        Select \(X^{(i^*)} \notin \overline Z\) which maximises \(%
            \min_{z^{(l)} \in \overline Z} \left\{%
                \text{Dens}\left(X^{(i)}\right) \times
                d\left(X^{i}, z^{(l)}\right)
            \right\}
        \)\;

        \(\overline Z \gets \overline Z \cup \left\{X^{(i^*)}\right\}\)
    }
\ealg%

\section{Matching games and the proposed method}\label{sec:method}

Both of the initialisation methods described in Section~\ref{subsec:inits} have
a greedy component. Cao's method essentially chooses the densest point that has
not already been chosen whilst forcing separation between the set of initial
modes. In the case of Huang's, however, the greediness only comes at the end
of the method, when the set of potential modes is replaced by a set of instances
in the dataset. Specifically, this means that in any practical implementation of
this method the order in which a set of potential modes is iterated over can
affect the set of initial modes. Thus, there is no guarantee of consistency.

The initialisation proposed in this work extends Huang's method to be
order-invariant in the final allocation --- thereby eliminating its greedy
component --- and provides a more intuitive starting point for the \(k\)-modes
algorithm. This is done by constructing and solving a matching game between the
set of potential modes and some subset of the data.

In general, matching games are defined by two sets (parties) of players in which
each player creates a preference list of at least some of the players in the
other party. The objective then is to find a `stable' mapping between the two
sets of players such that no pair of players is (rationally) unhappy with their
matching. Algorithms to `solve' --- i.e.\ find stable matchings to --- instances
of matching games are often structured to be party-oriented and aim to maximise
some form of social or party-based optimality~\cite{%
    Erdil2017,Fuku2006,Gale1962,Iwama2016,Kwanashie2015,Manlove2002%
}.

The particular constraints of this case --- where the \(k\) potential modes must
be allocated to a nearby unique data point --- mirror those of the so-called
Hospital-Resident Assignment Problem (HR). This problem gets its name from the
real-world problem of fairly allocating medical students to hospital posts.  A
resident-optimal algorithm for solving HR was presented in~\cite{Gale1962} and
was adapted in~\cite{Roth1984} to take advantage of the structure of the game.
This adapted algorithm is given in Algorithm~\ref{alg:hospital_resident}. A
practical implementation of this algorithm has been implemented in Python as
part of the \texttt{matching} library~\cite{Matching1.1} and is used in the
implementation of the proposed method for Section~\ref{sec:results}.

The game used to model HR, its matchings, and its notion of stability are
defined in Definitions~\ref{def:game}---\ref{def:blocking}. A summary of these
definitions in the context of the proposed \(k\)-modes initialisation is given
in Table~\ref{tab:components} before a formal statement of the proposed method
in Algorithm~\ref{alg:proposed_method}.

\begin{definition}\label{def:game}
    Consider two distinct sets \(R, H\) and refer to them residents and
    hospitals. Each \(h \in H\) has a capacity \(c_h \in \mathbb{N}\) associated
    with them. Each player \(r \in R\) and \(h \in H\) has associated 
    with it a strict preference list of the other set's elements such that:
    \begin{itemize}
        \item Each \(r \in R\) ranks a non-empty subset of \(H\), denoted by
            \(f(r)\).
        \item Each \(h \in H\) ranks all and only those residents that have
            ranked it, i.e.\ the preference list of \(h\), denoted \(g(h)\), is
            a permutation of the set
            \(\left\{r \in R \ | \ h \in f(r)\right\}\). If no such residents
            exist, \(h\) is removed from \(H\).
    \end{itemize}

    This construction of residents, hospitals, capacities and preference lists
    is called a \emph{game} and is denoted by \((R, H)\).
\end{definition}

\begin{definition}\label{def:matching}
    Consider a game \((R, H)\). A \emph{matching} \(M\) is any mapping between
    \(R\) and \(H\). If a pair \((r, h) \in R \times H\) are matched in \(M\)
    then this relationship is denoted \(M(r) = h\) and \(r \in M^{-1}(h)\).

    A matching is only considered \emph{valid} if all of the following hold for
    all \(r \in R, h \in H\):
    \begin{itemize}
        \item If \(r\) is matched then \(M(r) \in f(r)\).
        \item If \(h\) has at least one match then \(M^{-1}(h) \subseteq g(h)\).
        \item \(h\) is not over-subscribed, i.e.\ \(\abs*{M^{-1}(h)} \leq c_h\).
    \end{itemize}

    A valid matching is considered \emph{stable} if it does not contain any
    blocking pairs.
\end{definition}

\begin{definition}\label{def:blocking}
    Consider a game \((R, H)\). Then a pair \((r, h) \in R \times H\) is said to
    \emph{block} a matching \(M\) if all of the following hold:
    \begin{itemize}
        \item There is mutual preference, i.e.\ \(r \in g(h)\) and \(h \in
            f(r)\).
        \item Either \(r\) is unmatched or they prefer \(h\) to \(M(r)\).
        \item Either \(h\) is under-subscribed or \(h\) prefers \(r\) to at
            least one resident in \(M^{-1}(h)\).
    \end{itemize}
\end{definition}

\begin{table}[htbp]
    \resizebox{\textwidth}{!}{%
    \begin{tabular}{lcr}
        \toprule%
        Object in \(k\)-modes initialisation & {} & Object in a matching game
        \\\midrule%
        Potential modes & {} & The set of residents
        \\
        Data points closest to potential modes & {} & The set of hospitals
        \\
        Similarity between a potential mode and a point & {} & Respective
        position in each other's preference lists
        \\
        The data point to replace a potential mode & {} & A pair in a matching
        \\\bottomrule
    \end{tabular}}
    \caption{A summary of the relationships between the components of the
             initialisation for \(k\)-modes and those in a matching game
             \((R, H)\).
    }\label{tab:components}
\end{table}

\balg%
    \caption{The hospital-resident algorithm
        (resident-optimal)}\label{alg:hospital_resident}
    \KwIn{a set of residents \(R\), a set of hospitals \(H\), a set of hospital
        capacities \(C\), two preference list functions \(f, g\)}
    \KwOut{a stable, resident-optimal mapping \(M\) between \(R\) and \(H\)}

    \For{\(h \in H\)}{%
        \(M^{-1}(h) \gets \emptyset\)
    }
    \While{There exists any unmatched \(r \in R\) with a non-empty preference
        list}{%
        Take any such resident \(r\) and their most preferred hospital \(h\)\;
        \(\textsc{MatchPair}(s, h)\)\;

        \If{\(\abs*{M^{-1}(h)} > c_h\)}{%
            Find their worst match \(r' \in M^{-1}(h)\)\;
            \(\textsc{UnmatchPair}(r', h)\)\;
        }
        \If{\(\abs*{M^{-1}(h)} = c_h\)}{%
            Find their worst match \(r' \in M^{-1}(h)\)\;
            \For{each successor \(s \in g(h)\) to \(r'\)}{%
                \(\textsc{DeletePair}(s, h)\)
            }
        }
    }
\ealg%

\balg%
    \caption{\textsc{MatchPair}}\label{alg:match}
    \KwIn{a resident \(r\), a hospital \(h\), a matching \(M\)}
    \KwOut{an updated matching \(M\)}

    \(M^{-1}(h) \gets M^{-1}(h) \cup \left\{r\right\}\)\;
\ealg%

\balg%
    \caption{\textsc{UnmatchPair}}\label{alg:unmatch}
    \KwIn{a resident \(r\), a hospital \(h\), a matching \(M\)}
    \KwOut{an updated matching \(M\)}

    \(M^{-1}(h) \gets M^{-1}(h) \setminus \left\{r\right\}\)\;
\ealg%

\balg%
    \caption{\textsc{DeletePair}}\label{alg:delete}
    \KwIn{a resident \(r\), a hospital \(h\)}
    \KwOut{updated preference lists}

    \(f(r) \gets f(r) \setminus \left\{h\right\}\)\;
    \(g(h) \gets g(h) \setminus \left\{r\right\}\)\;
\ealg%

\balg%
    \caption{The proposed initialisation method}\label{alg:proposed_method}
    \KwIn{a dataset \(\mathcal{X} \subset \mathcal{A}\), a number of modes to
        find \(k\)}
    \KwOut{a set of \(k\) initial modes \(\overline Z\)}

    \(\overline Z \gets \emptyset\)\;
    \(H \gets \emptyset\)\;
    \(R \gets \textsc{SamplePotentialModes}\left(\mathcal{X}\right)\)\;

    \For{\(r \in R\)}{%
        Find the set of \(k\) data points \(H_r \subset \mathcal{X}\) that 
        are the least dissimilar to \(r\)\;
        Arrange \(H_r\) into descending order of similarity with respect to
        \(r\), denoted by \(H_r^*\)\;
        \(H \gets H \cup H_r\)\;
        \(f(r) \gets H_r^*\)\;
    }

    \For{\(h \in H\)}{%
        \(c_h \gets 1\)\;
        Sort \(R\) into descending order of similarity with respect to \(h\),
        denoted by \(R^*\)\;
        \(g(h) \gets R^*\)
    }

    Solve the matching game defined by \((R, H)\) to obtain a matching \(M\)\;
    \For{\(r \in R\)}{%
        \(\overline Z \gets \overline Z \cup \left\{M(r)\right\}\)
    }
\ealg%

\section{Experimental results}\label{sec:results}
\graphicspath{{./img/}}

To give comparative results on the quality of the initialisation processes
considered in this work, four well-known, categorical, labelled datasets ---
breast cancer, mushroom, nursery, and soybean (large) --- will be clustered by
the \(k\)-modes algorithm with each of the initialisation processes. These
datasets have been chosen to fall in line with the established literature, and
for their relative sizes and complexities. Each dataset is openly available
under the UCI Machine Learning Repository~\cite{Dua2019}, and their
characteristics are summarised in Table~\ref{tab:dataset_summary}. For the
purposes of this analysis, incomplete instances (i.e.\ where data is missing)
are excluded and the remaining dataset characteristics are reported as
`adjusted'.

\begin{table}[htbp]
    \resizebox{\textwidth}{!}{%
\begin{tabular}{lrrrrrrr}
\toprule
{} &      N &   m &  No. classes &  Missing values &  Adjusted N &  Adjusted no. classes &  No. clusters found \\
\midrule
Breast cancer &    699 &  10 &            2 &            True &         683 &                     2 &                   8 \\
Mushroom      &   8124 &  22 &            2 &            True &        5644 &                     2 &                  17 \\
Nursery       &  12960 &   8 &            5 &           False &       12960 &                     5 &                  23 \\
Soybean       &    307 &  35 &           19 &            True &         266 &                    15 &                   8 \\
\bottomrule
\end{tabular}

    }\caption{A summary of the benchmark datasets.}\label{tab:dataset_summary}
\end{table}

All of the source code used to produce the results and data in this analysis ---
including the datasets investigated in Section~\ref{subsec:artificial} --- are
archived at
DOI~\href{https://doi.org/10.5281/zenodo.3639282}{10.5281/zenodo.3639282}. In
addition to this, the implementation of the \(k\)-modes algorithm and its
initialisations is available under
DOI~\href{https://doi.org/10.5281/zenodo.3638035}{10.5281/zenodo.3638035}.

This analysis does not consider evaluative metrics related to classification
such as accuracy, recall or precision as is commonly done~\cite{%
    Arthur2007,Cao2009,Cao2012,Huang1998,%
    Ng2007,Olaode2014,Schaeffer2007,Sharma2015%
}. Instead, only internal measures are considered such as the cost function
defined in~\eqref{eq:cost}. This metric is label-invariant and its values are
comparable across the different initialisation methods. Furthermore, the effect
of each initialisation method on the initial and final clusterings can be
captured with the cost function. An additional, and often useful, metric is the
silhouette coefficient. This measures the ratio between the intra-cluster
cohesion and inter-cluster separation of a particular clustering. Therefore, it
could be used in a similar way to reveal the effect of each initialisation
method at the beginning and end of a run of \(k\)-modes. Unfortunately, this
metric loses its intuition under the distance measure employed here and is
omitted. The remaining performance measures used are the number of iterations
for the \(k\)-modes algorithm to terminate and the time taken to terminate in
seconds.

The final piece of information required in this analysis is a choice for \(k\)
for each dataset. An immediate choice is the number of classes that are present
in a dataset but this is not necessarily an appropriate choice since the classes
may not be representative of true clusters~\cite{Memoli2011}. However, this
analysis will consider this case as there may be practical reasons to limit the
value of \(k\). The other strategy for choosing \(k\) considered in this work
uses the knee point detection algorithm introduced in~\cite{Satopaa2011}. This
strategy was chosen over other popular methods such as the `elbow' method as its
results are definitive.

The knee point detection algorithm was employed over values of \(k\) from 2 up
to \(\lfloor\sqrt N\rfloor\) for each dataset. The number of clusters determined
by this strategy is reported in the final column of
Table~\ref{tab:dataset_summary}.

\subsection{Using knee point detection algorithm for \(k\)}\label{subsec:knee}

Tables~\ref{tab:breast_cancer_knee}---\ref{tab:soybean_knee}
summarise the results of each initialisation method on the benchmark datasets
where the number of clusters has been determined by the knee point detection
algorithm. Each column shows the mean value of each metric and its standard
250
repetitions of the \(k\)-modes algorithm.

\begin{table}[htbp]
    \centering
    \resizebox{\tablewidth}{!}{%
\begin{tabular}{lllll}
\toprule
{} &       Initial cost &        Final cost & No. iterations &          Time \\
\midrule
Cao      &    3118.00 (0.000) &   2774.00 (0.000) &   4.00 (0.000) &  0.30 (0.012) \\
Huang    &  2856.50 (104.245) &  2748.83 (64.514) &   2.68 (0.817) &  0.22 (0.046) \\
Matching &  2870.11 (101.869) &  2752.59 (52.387) &   2.72 (0.760) &  0.16 (0.021) \\
\bottomrule
\end{tabular}

    }
    \captionof{table}{Summative metric results for the breast cancer dataset
    with \(k=8\).}\label{tab:breast_cancer_knee}\vspace{20pt}

    \resizebox{\tablewidth}{!}{%
\begin{tabular}{lllll}
\toprule
{} &         Initial cost &          Final cost & No. iterations &          Time \\
\midrule
Cao      &     20381.00 (0.000) &    20376.00 (0.000) &   2.00 (0.000) &  4.68 (0.205) \\
Huang    &  23027.24 (1209.753) &  21869.06 (747.766) &   2.90 (0.934) &  5.11 (1.138) \\
Matching &  23279.36 (1498.324) &  21855.50 (751.641) &   3.02 (0.936) &  2.77 (0.325) \\
\bottomrule
\end{tabular}

    }
    \captionof{table}{Summative metric results for the mushroom dataset with
    \(k=17\).}\label{tab:mushroom_knee}\vspace{20pt}

    \resizebox{\tablewidth}{!}{%
\begin{tabular}{lllll}
\toprule
{} &        Initial cost &          Final cost & No. iterations &          Time \\
\midrule
Cao      &    35544.00 (0.000) &    35544.00 (0.000) &   1.00 (0.000) &  4.98 (0.152) \\
Huang    &  37535.06 (372.596) &  37535.06 (372.596) &   1.00 (0.000) &  3.58 (0.121) \\
Matching &  37484.29 (327.467) &  37484.29 (327.467) &   1.00 (0.000) &  3.14 (0.141) \\
\bottomrule
\end{tabular}

    }
    \captionof{table}{Summative metric results for the nursery dataset with
    \(k=23\).}\label{tab:nursery_knee}\vspace{20pt}

    \resizebox{\tablewidth}{!}{%
\begin{tabular}{lllll}
\toprule
{} &      Initial cost &        Final cost & No. iterations &          Time \\
\midrule
Cao      &   1654.00 (0.000) &   1585.00 (0.000) &   4.00 (0.000) &  0.28 (0.014) \\
Huang    &  1829.31 (92.308) &  1708.55 (69.740) &   3.58 (1.019) &  0.28 (0.063) \\
Matching &  1827.76 (86.852) &  1711.49 (73.319) &   3.42 (0.963) &  0.17 (0.022) \\
\bottomrule
\end{tabular}

    }
    \captionof{table}{Summative metric results for the soybean dataset with
    \(k=8\).}\label{tab:soybean_knee}
\end{table}

By examining these tables it would seem that the proposed method and Huang's
method are comparable across the board --- although the proposed method is
faster despite taking more iterations in general which may relate to a more
intuitive initialisation. More importantly though, it appears that Cao's method
performs the best out of the three initialisation methods: in terms of initial
and final costs Cao's method improves, on average, by roughly 10 percent against
the next best method for the three datasets that it succeeds with; the number of
iterations is comparable; and the computation time is substantially less than
the other two considering it is a deterministic method and need only be run once
to achieve this performance.

\begin{figure}
    \begin{subfigure}{.5\textwidth}
        \includegraphics[width=\linewidth]{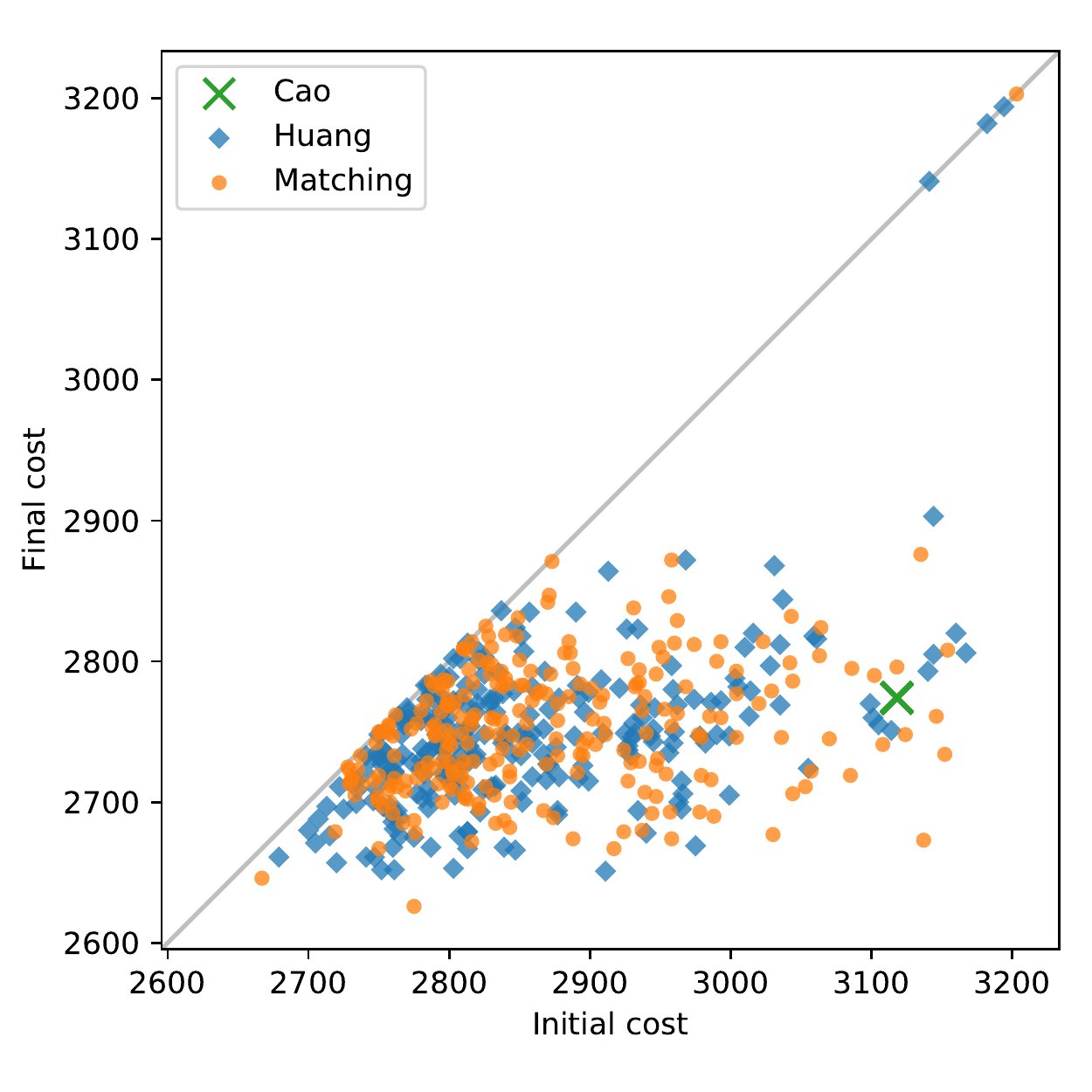}
        \caption{Scatter plot of initial and final costs.}
    \end{subfigure}
    \hfill%
    \begin{subfigure}{.5\textwidth}
        \includegraphics[width=\linewidth]{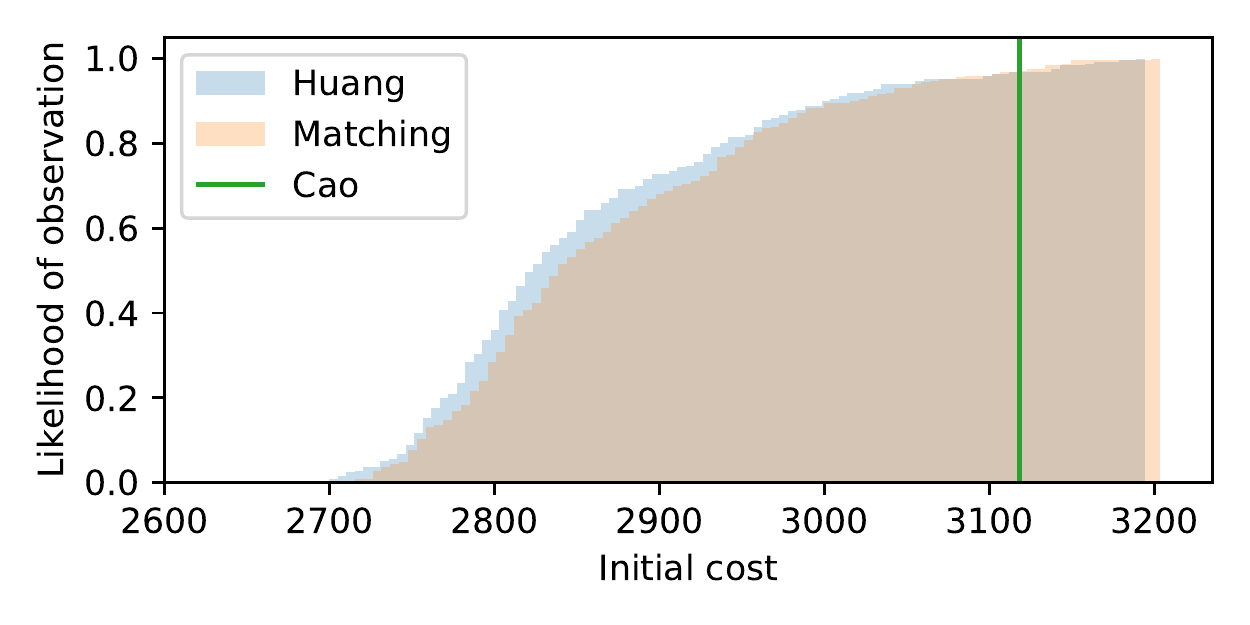}

        \includegraphics[width=\linewidth]{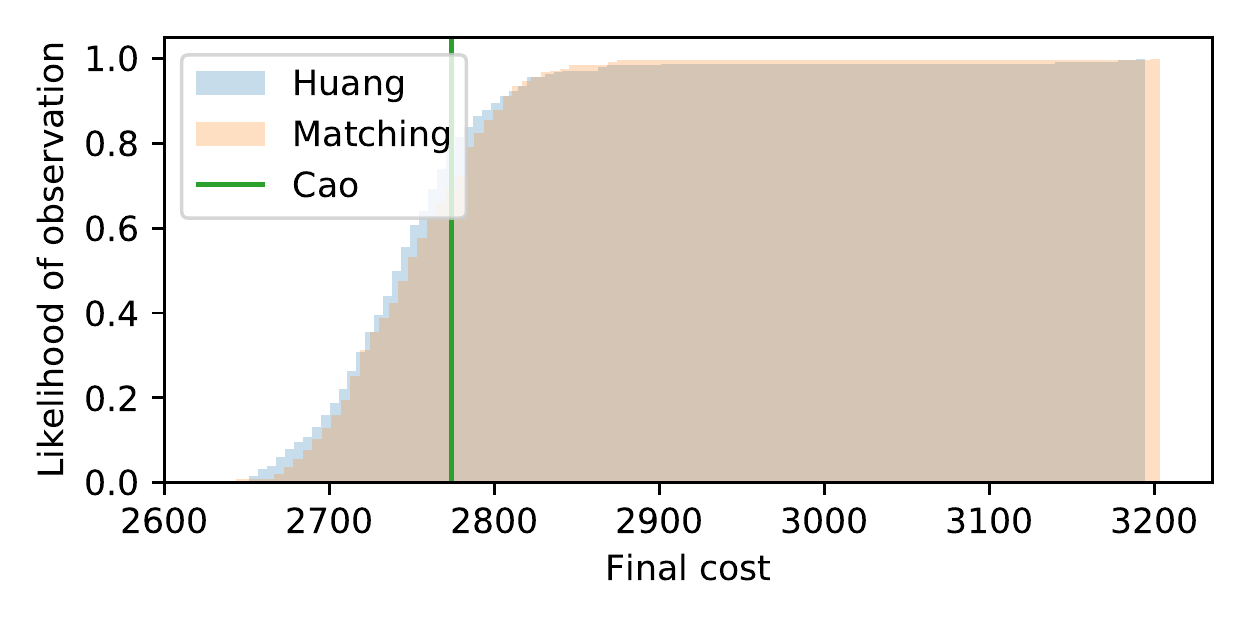}
        \caption{Empirical CDF plots for initial (top) and final (bottom)
                 costs.}
    \end{subfigure}
    \caption{Summative plots for the breast cancer dataset with \(k=8\).}%
    \label{fig:breast_cancer_knee}
\end{figure}

\begin{figure}
    \begin{subfigure}{.5\textwidth}
        \includegraphics[width=\linewidth]{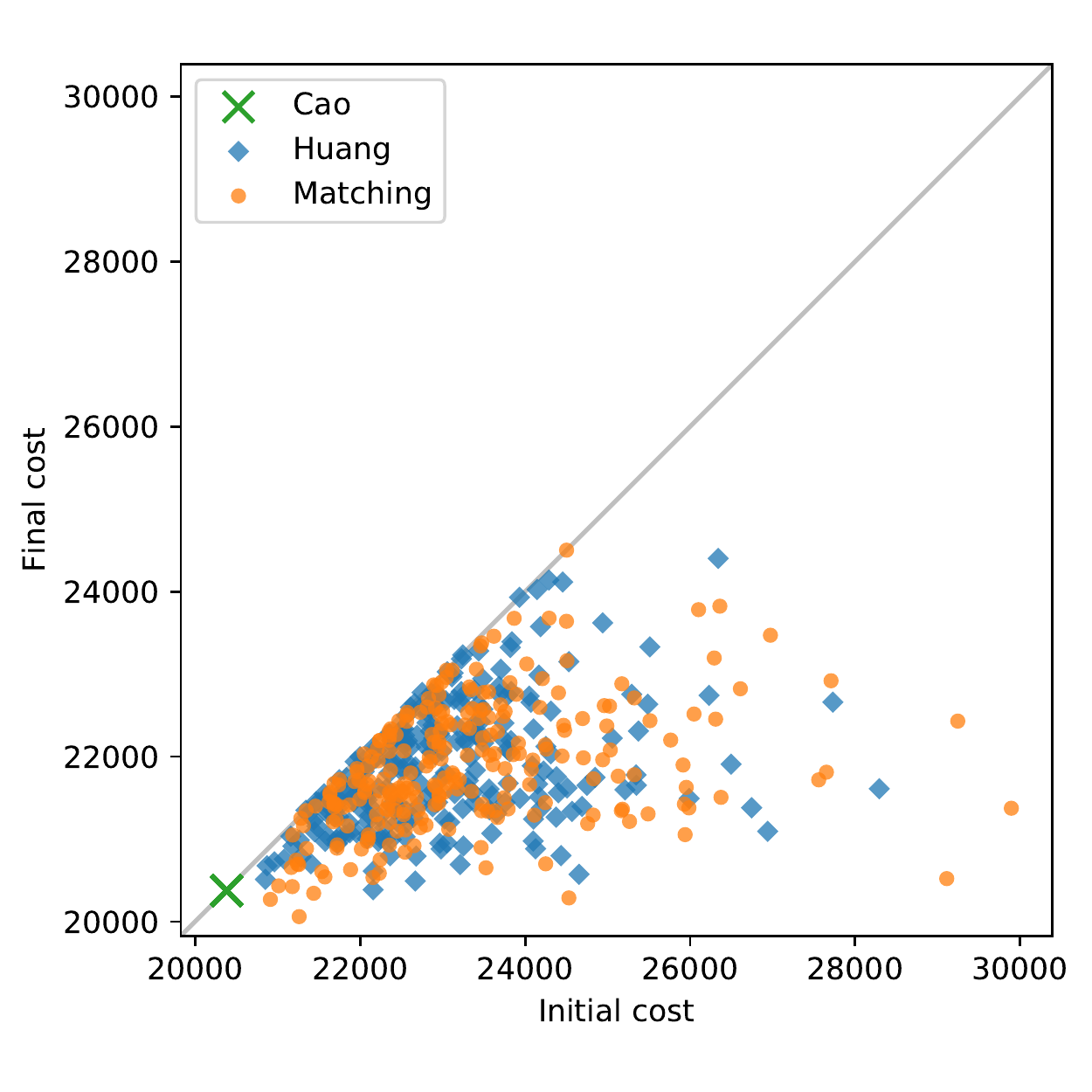}
        \caption{Scatter plot of initial and final costs.}
    \end{subfigure}
    \hfill%
    \begin{subfigure}{.5\textwidth}
        \includegraphics[width=\linewidth]{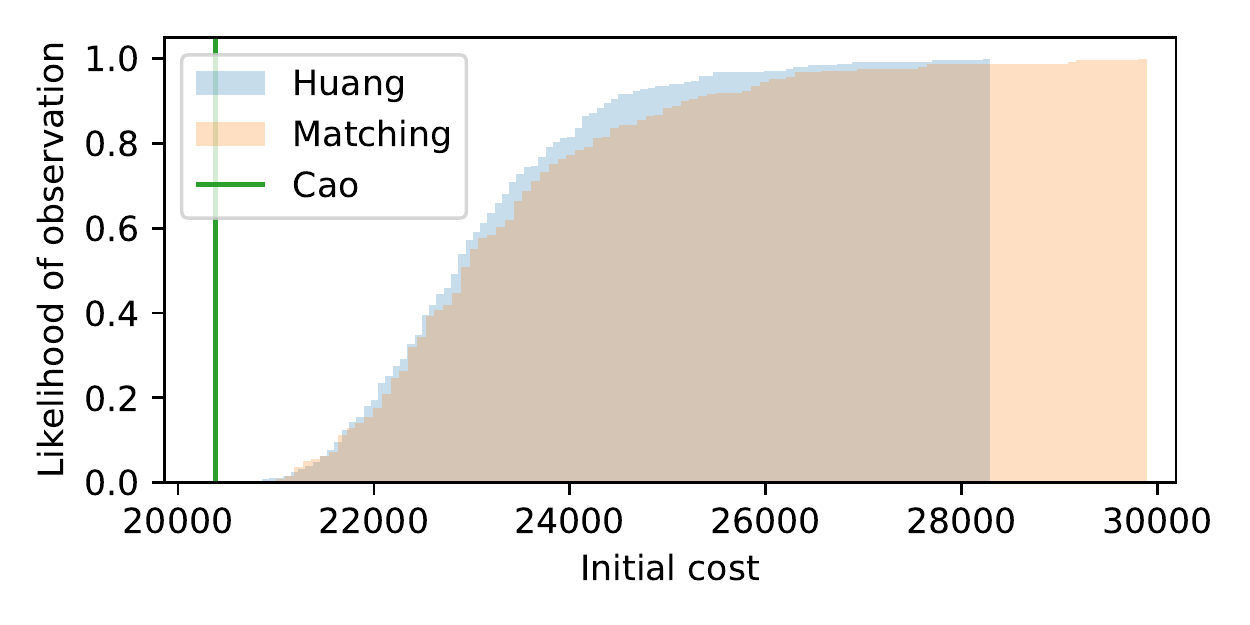}

        \includegraphics[width=\linewidth]{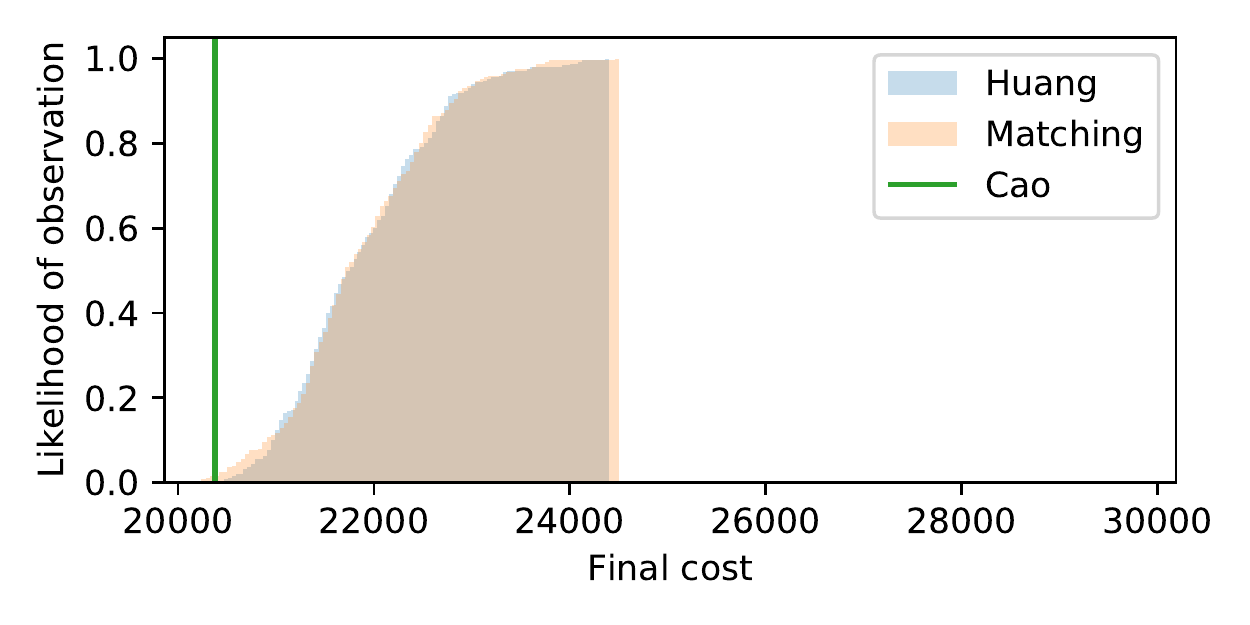}
        \caption{Empirical CDF plots for initial (top) and final (bottom)
                 costs.}
    \end{subfigure}
    \caption{Summative plots for the mushroom dataset with \(k=17\).}%
    \label{fig:mushroom_knee}
\end{figure}

\begin{figure}
    \begin{subfigure}{.5\textwidth}
        \includegraphics[width=\linewidth]{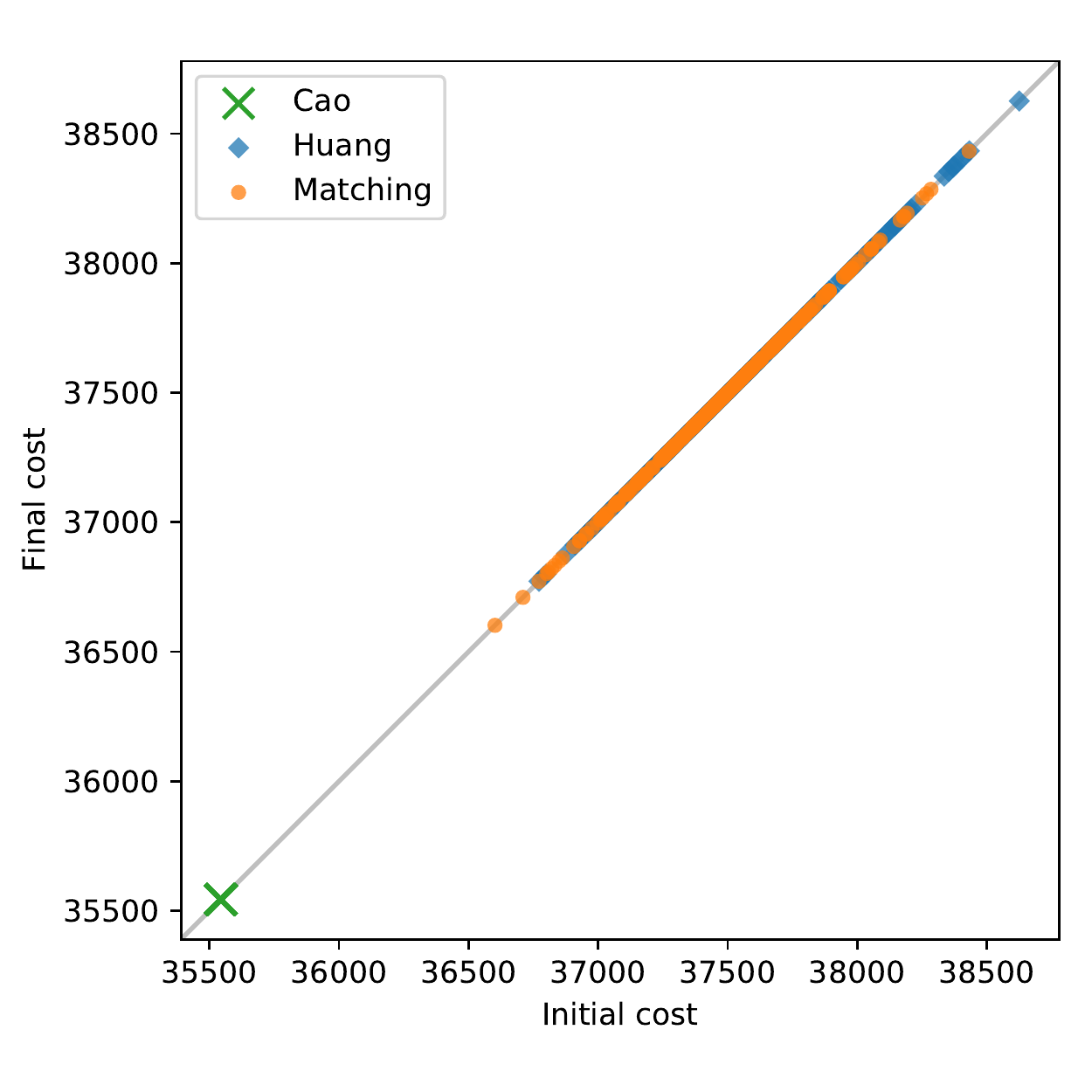}
        \caption{Scatter plot of initial and final costs.}
    \end{subfigure}
    \hfill%
    \begin{subfigure}{.5\textwidth}
        \includegraphics[width=\linewidth]{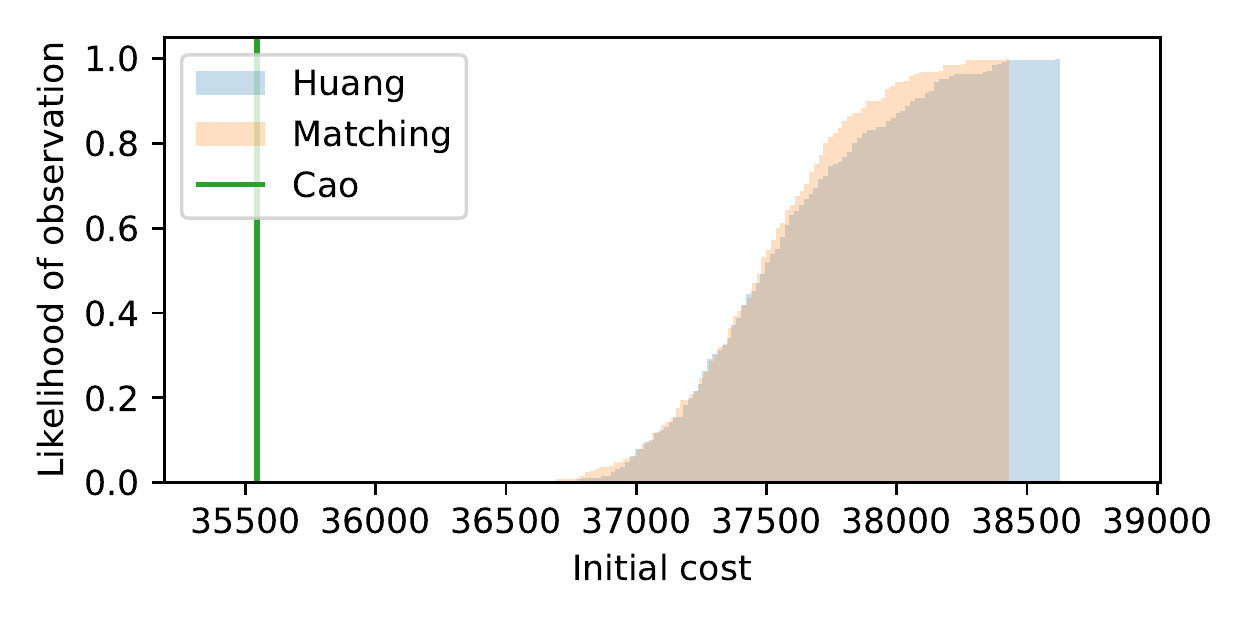}

        \includegraphics[width=\linewidth]{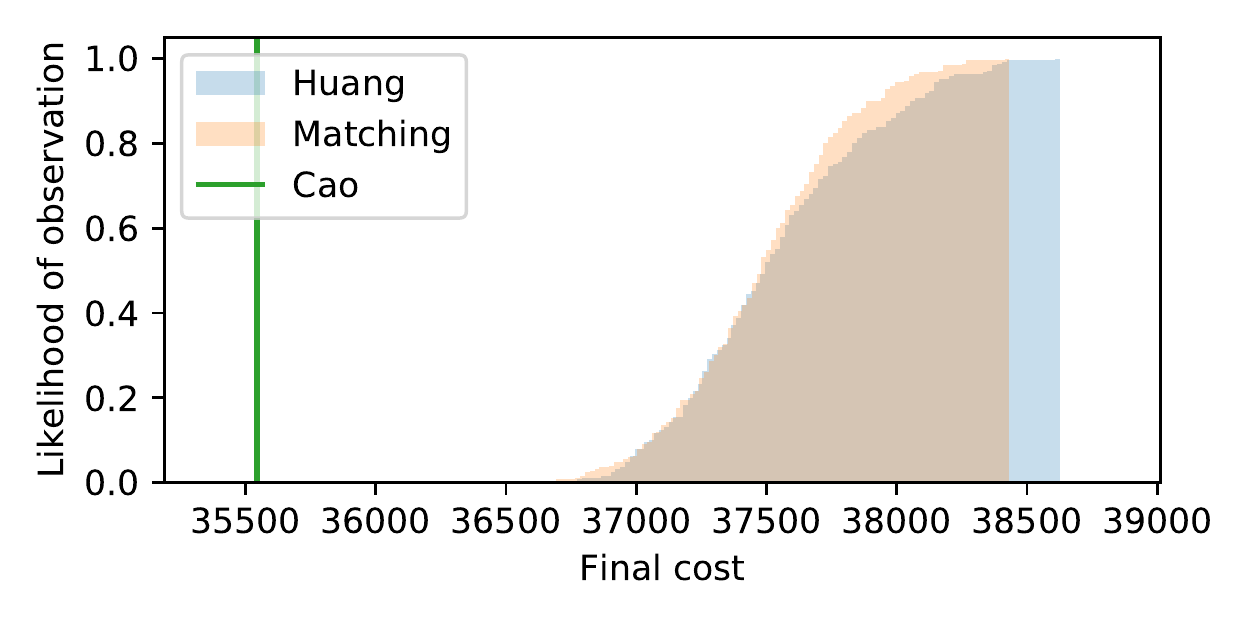}
        \caption{Empirical CDF plots for initial (top) and final (bottom)
                 costs.}
    \end{subfigure}
    \caption{Summative plots for the nursery dataset with \(k=23\).}%
    \label{fig:nursery_knee}
\end{figure}

\begin{figure}
    \begin{subfigure}{.5\textwidth}
        \includegraphics[width=\linewidth]{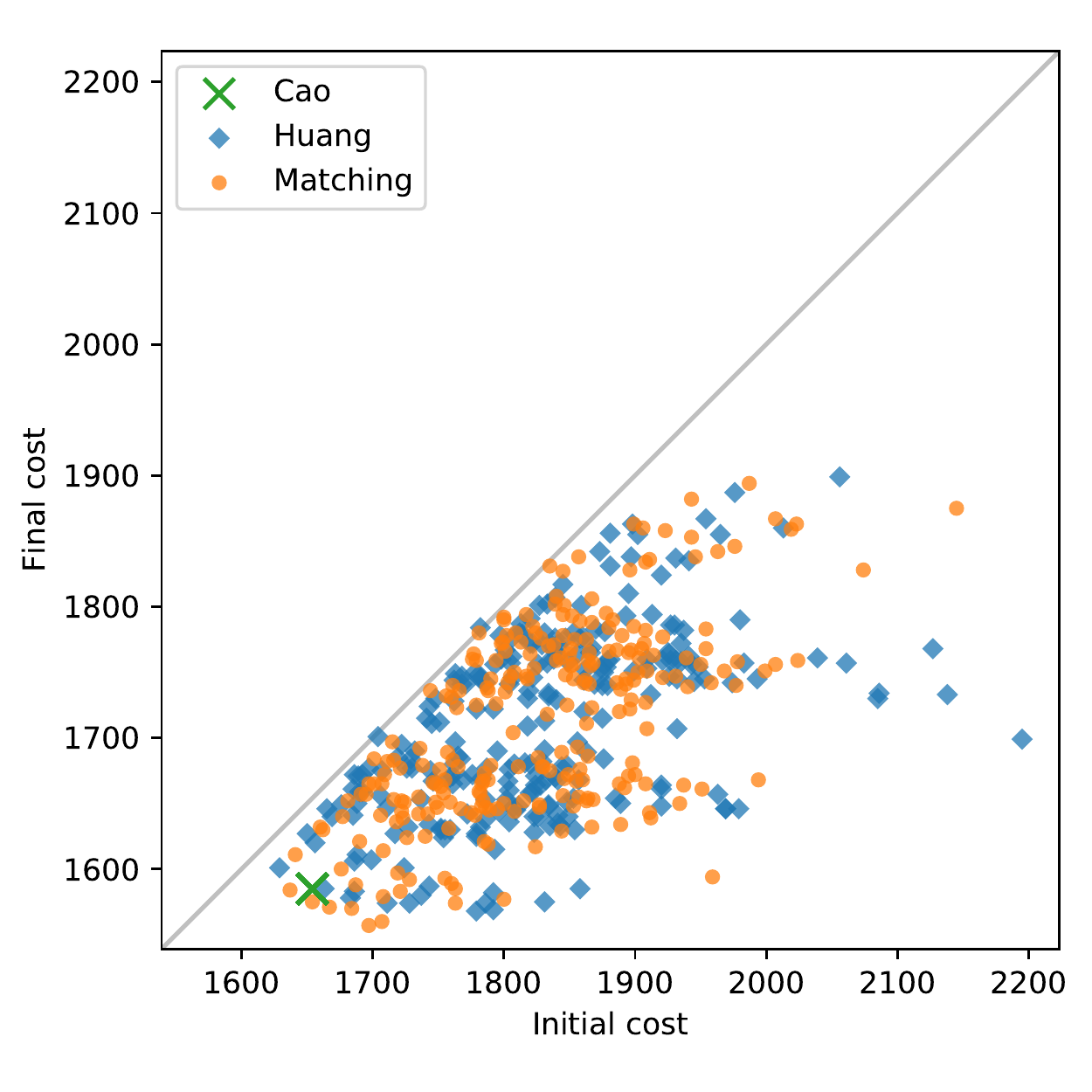}
        \caption{Scatter plot of initial and final costs.}
    \end{subfigure}
    \hfill%
    \begin{subfigure}{.5\textwidth}
        \includegraphics[width=\linewidth]{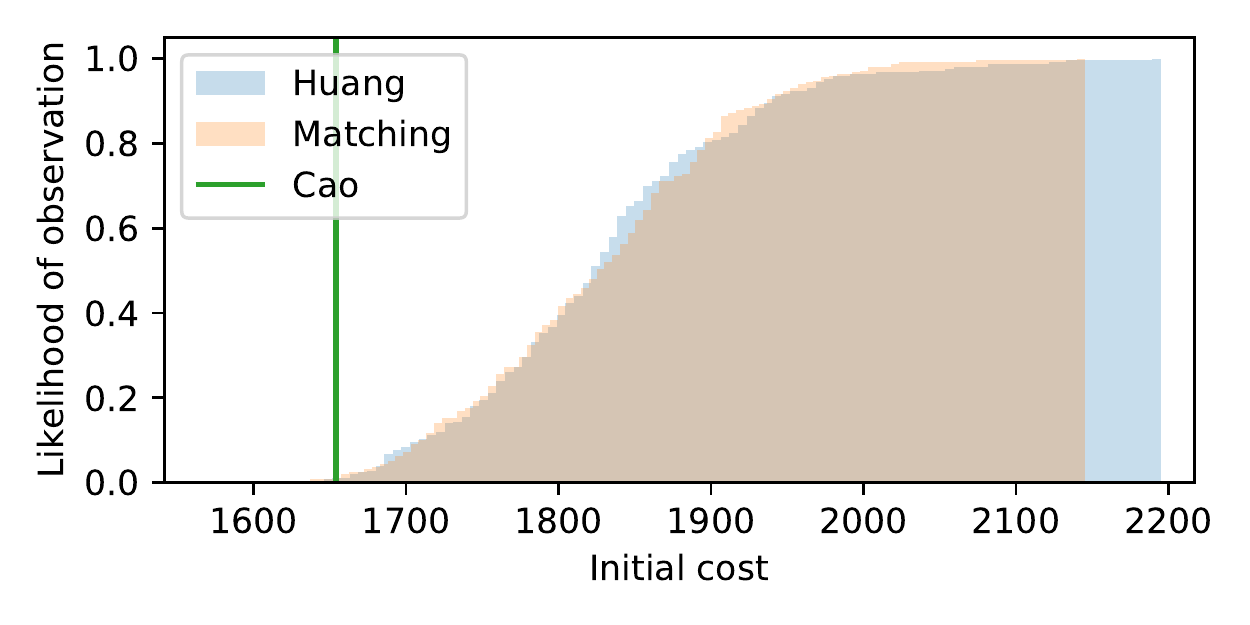}

        \includegraphics[width=\linewidth]{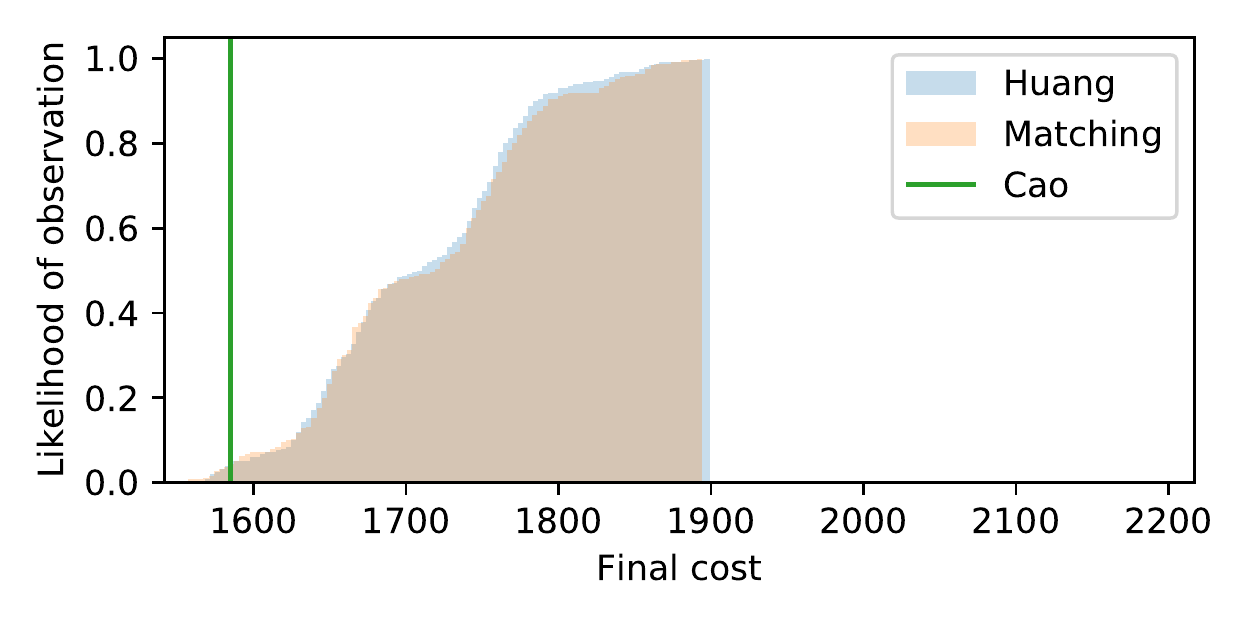}
        \caption{Empirical CDF plots for initial (top) and final (bottom)
                 costs.}
    \end{subfigure}
    \caption{Summative plots for the soybean dataset with \(k=8\).}%
    \label{fig:soybean_knee}
\end{figure}

However, in the \(k\)-means paradigm, a particular clustering is selected based
on it having the minimum final cost over a number of runs of the algorithm ---
not the mean --- and while Cao's method is very reliable, in that there is no
variation at all, it does not always produce the best clustering possible. There
is a trade-off to be made between computational time and performance here. In
order to gain more insight into the performance of each method, less granular
analysis is required.
Figures~\ref{fig:breast_cancer_knee}---\ref{fig:soybean_knee} display the
cost function results for each dataset in the form of a scatter plot and two
empirical cumulative density function (CDF) plots, highlighting the breadth and
depth of the behaviours exhibited by each initialisation method.

Looking at Figure~\ref{fig:breast_cancer_knee} it is clear that in terms of
final cost Cao's method is middling when compared to the other methods. This
was apparent from Table~\ref{tab:breast_cancer_knee} and, indeed, Huang's and
the proposed method are both very comparable when looking at the main body of
the results. However, since the criterion for the best clustering (in practical
terms) is having the minimum final cost, it is evident that the proposed method
is superior; that the method produces clusterings with a larger cost
range (indicated by the trailing right-hand side of each CDF plot) is irrelevant
for the same reason.

This pattern of largely similar behaviour between Huang's and the proposed
method is apparent in each of the figures here, and in each case the proposed
method outperforms Huang's. In fact, in all cases except for the nursery
dataset, the proposed method achieves the lowest final cost of all the methods
and, as such, performs the best in practical terms on these particular datasets.

In the case of the nursery dataset, Cao's method is unquestionably the best
performing initialisation method. It should be noted that none of the
methods were able to find an initial clustering that could be improved on, and
that this dataset exactly describes the entire attribute space in which it
exists. This property could be why the other methods fall behind Cao's so
decisively in that Cao's method is able to definitively choose the \(k\) most
dense-whilst-separated points from the attribute space as the initial cluster
centres whereas the other two methods are in essence randomly sampling from this
space. That each initial solution in these repetitions is locally optimal
remains a mystery.

\subsection{Using number of classes for \(k\)}\label{subsec:nclasses}

As is discussed above, the often automatic choice for \(k\) is the number of
classes present in the data; this subsection repeats the analysis from the
subsection above but with this traditional choice for \(k\).
Tables~\ref{tab:breast_cancer_nclasses}---\ref{tab:soybean_nclasses}
contain the analogous summaries of each initialisation method's performance on
the benchmark datasets over the same number of repetitions.

\begin{table}[htbp]
    \centering
    \resizebox{\tablewidth}{!}{%
\begin{tabular}{lllll}
\toprule
{} &       Initial cost &         Final cost & No. iterations &          Time \\
\midrule
Cao      &    3315.00 (0.000) &    3172.00 (0.000) &   2.00 (0.000) &  0.13 (0.005) \\
Huang    &  3393.80 (120.772) &  3348.51 (144.849) &   1.54 (0.653) &  0.10 (0.024) \\
Matching &  3406.73 (111.686) &  3355.56 (144.621) &   1.61 (0.638) &  0.09 (0.018) \\
\bottomrule
\end{tabular}

    }
    \captionof{table}{Summative metric results for the breast cancer dataset
    with \(k=2\).}\label{tab:breast_cancer_nclasses}\vspace{20pt}

    \resizebox{\tablewidth}{!}{%
\begin{tabular}{lllll}
\toprule
{} &         Initial cost &           Final cost & No. iterations &          Time \\
\midrule
Cao      &     37662.00 (0.000) &     37662.00 (0.000) &   1.00 (0.000) &  0.94 (0.035) \\
Huang    &  41974.07 (2393.889) &  39226.25 (2483.933) &   3.11 (1.430) &  1.92 (0.679) \\
Matching &  42175.54 (2520.163) &  39617.53 (2637.574) &   3.03 (1.439) &  1.38 (0.491) \\
\bottomrule
\end{tabular}

    }
    \captionof{table}{Summative metric results for the mushroom dataset with
    \(k=2\).}\label{tab:mushroom_nclasses}\vspace{20pt}

    \resizebox{\tablewidth}{!}{%
\begin{tabular}{lllll}
\toprule
{} &        Initial cost &          Final cost & No. iterations &          Time \\
\midrule
Cao      &    49060.00 (0.000) &    49060.00 (0.000) &   1.00 (0.000) &  1.80 (0.090) \\
Huang    &  51229.45 (902.503) &  51229.45 (902.503) &   1.00 (0.000) &  1.72 (0.116) \\
Matching &  51107.52 (910.258) &  51101.95 (903.525) &   1.00 (0.063) &  1.37 (0.128) \\
\bottomrule
\end{tabular}

    }
    \captionof{table}{Summative metric results for the nursery dataset with
    \(k=5\).}\label{tab:nursery_nclasses}\vspace{20pt}

    \resizebox{\tablewidth}{!}{%
\begin{tabular}{lllll}
\toprule
{} &      Initial cost &        Final cost & No. iterations &          Time \\
\midrule
Cao      &   1364.00 (0.000) &   1314.00 (0.000) &   2.00 (0.000) &  0.33 (0.009) \\
Huang    &  1588.89 (83.682) &  1446.22 (59.844) &   4.02 (1.081) &  0.45 (0.085) \\
Matching &  1582.56 (87.418) &  1447.08 (60.154) &   4.01 (1.128) &  0.24 (0.025) \\
\bottomrule
\end{tabular}

    }
    \captionof{table}{Summative metric results for the soybean dataset with
    \(k=15\).}\label{tab:soybean_nclasses}
\end{table}

An immediate comparison to the previous tables is that for all datasets bar the
soybean dataset, the mean costs are significantly higher and the computation
times are lower. These effects come directly from the choice of \(k\) in that
higher values of \(k\) will require more checks (and thus computational time)
but will typically lead to more homogeneous clusters, reducing their
within-cluster dissimilarity and therefore cost.

\begin{figure}
    \begin{subfigure}{.5\textwidth}
        \includegraphics[width=\linewidth]{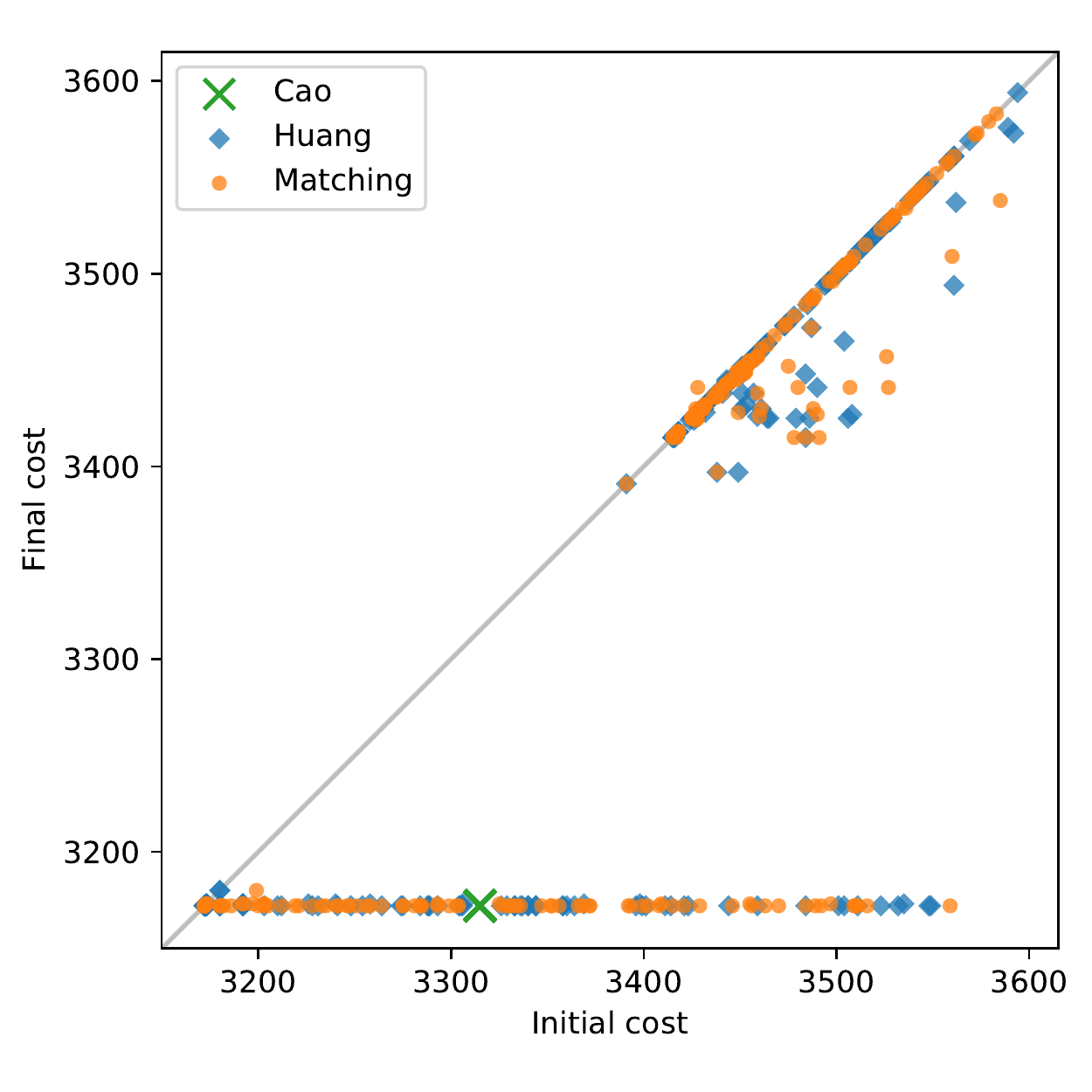}
        \caption{Scatter plot of initial and final costs.}
    \end{subfigure}
    \hfill%
    \begin{subfigure}{.5\textwidth}
        \includegraphics[width=\linewidth]{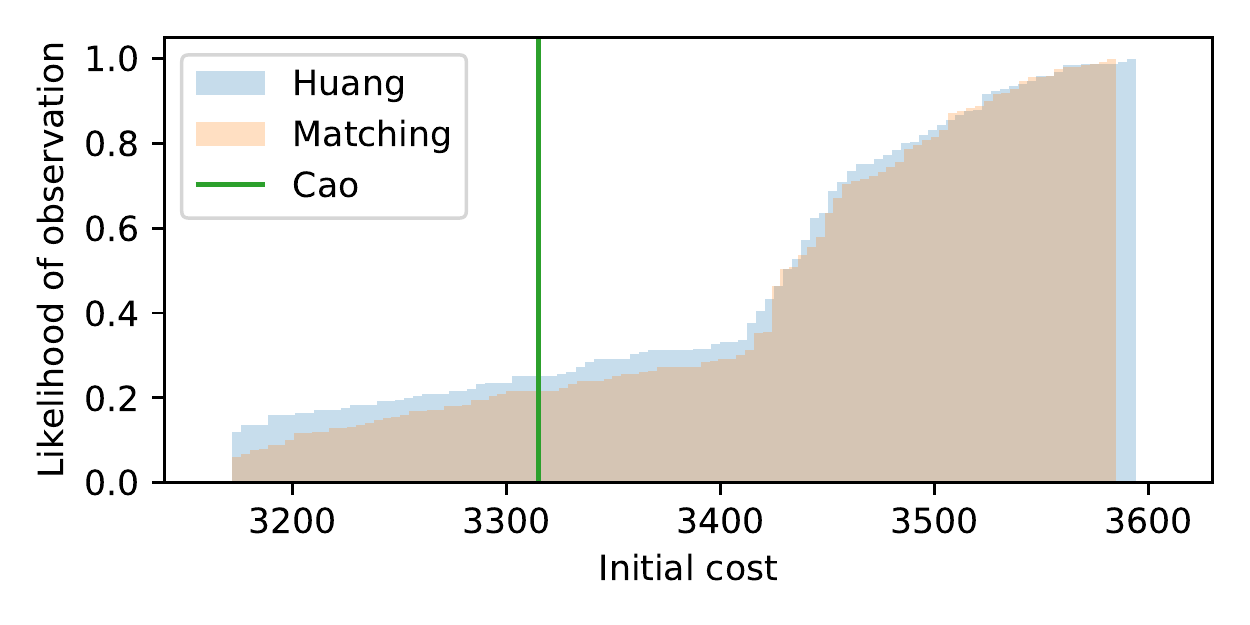}

        \includegraphics[width=\linewidth]{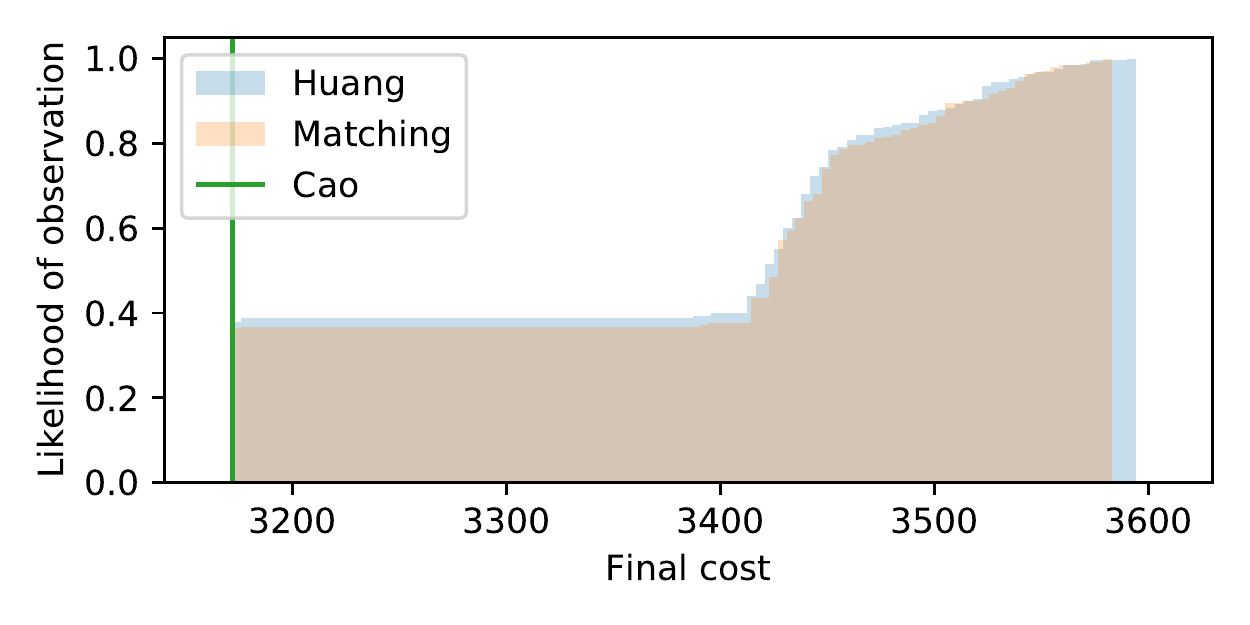}
        \caption{Empirical CDF plots for initial (top) and final (bottom)
                 costs.}
    \end{subfigure}
    \caption{Summative plots for the breast cancer dataset with \(k=2\).}%
    \label{fig:breast_cancer_nclasses}
\end{figure}

\begin{figure}
    \begin{subfigure}{.5\textwidth}
        \includegraphics[width=\linewidth]{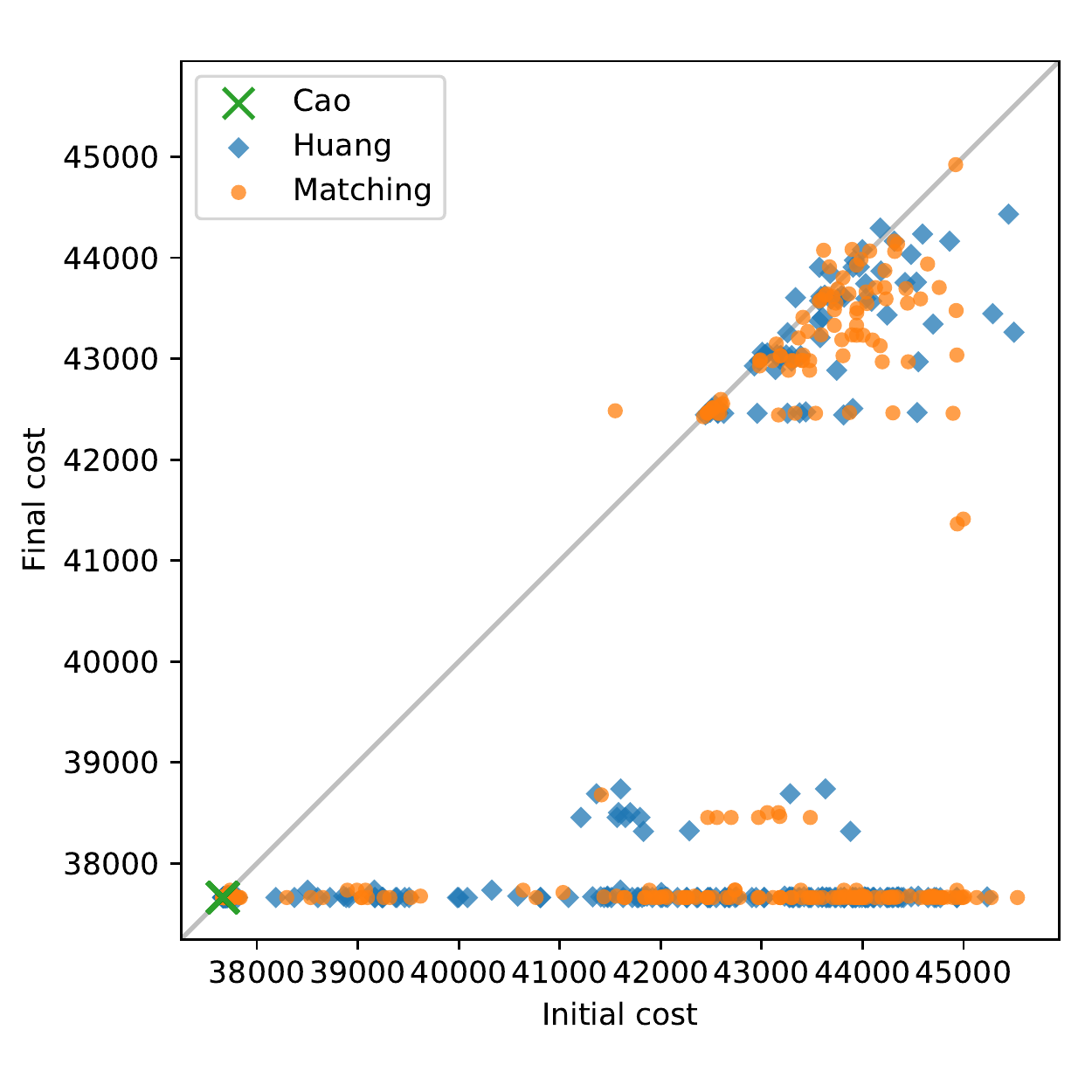}
        \caption{Scatter plot of initial and final costs.}
    \end{subfigure}
    \hfill%
    \begin{subfigure}{.5\textwidth}
        \includegraphics[width=\linewidth]{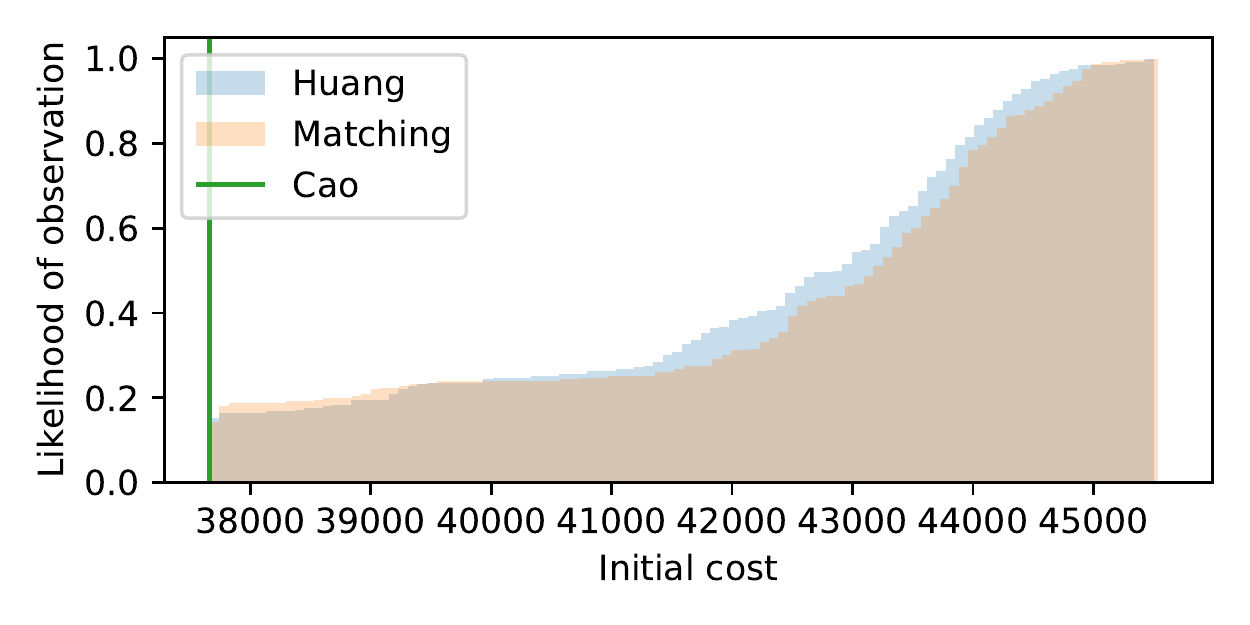}

        \includegraphics[width=\linewidth]{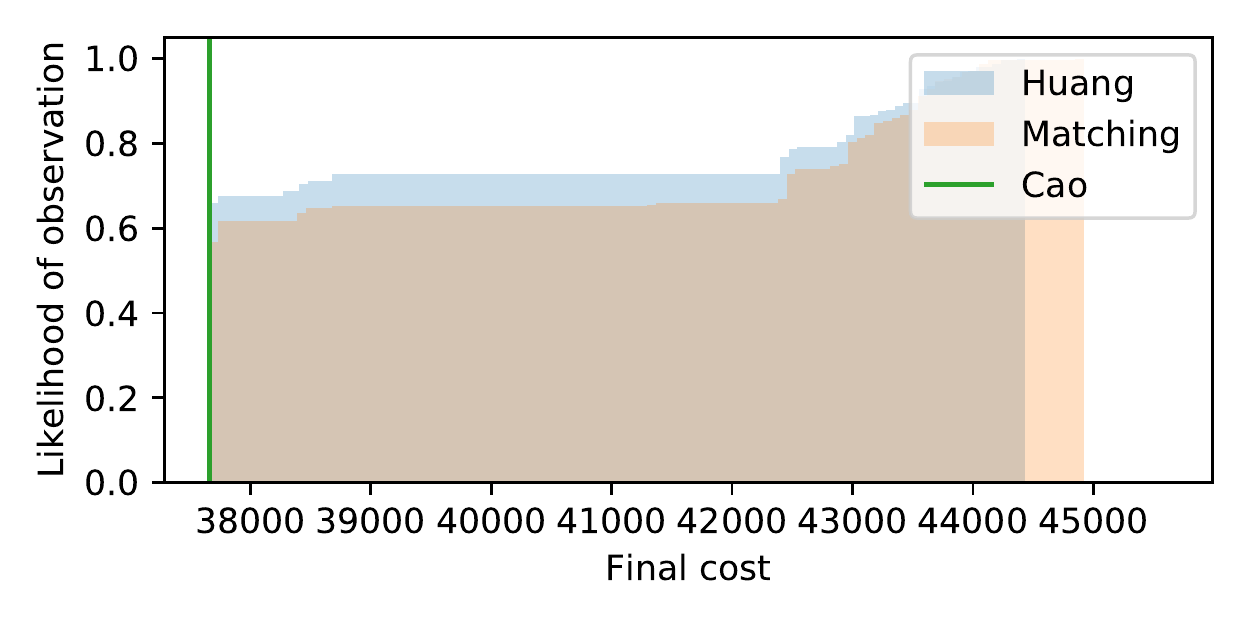}
        \caption{Empirical CDF plots for initial (top) and final (bottom)
                 costs.}
    \end{subfigure}
    \caption{Summative plots for the mushroom dataset with \(k=2\).}%
    \label{fig:mushroom_nclasses}
\end{figure}

\begin{figure}
    \begin{subfigure}{.5\textwidth}
        \includegraphics[width=\linewidth]{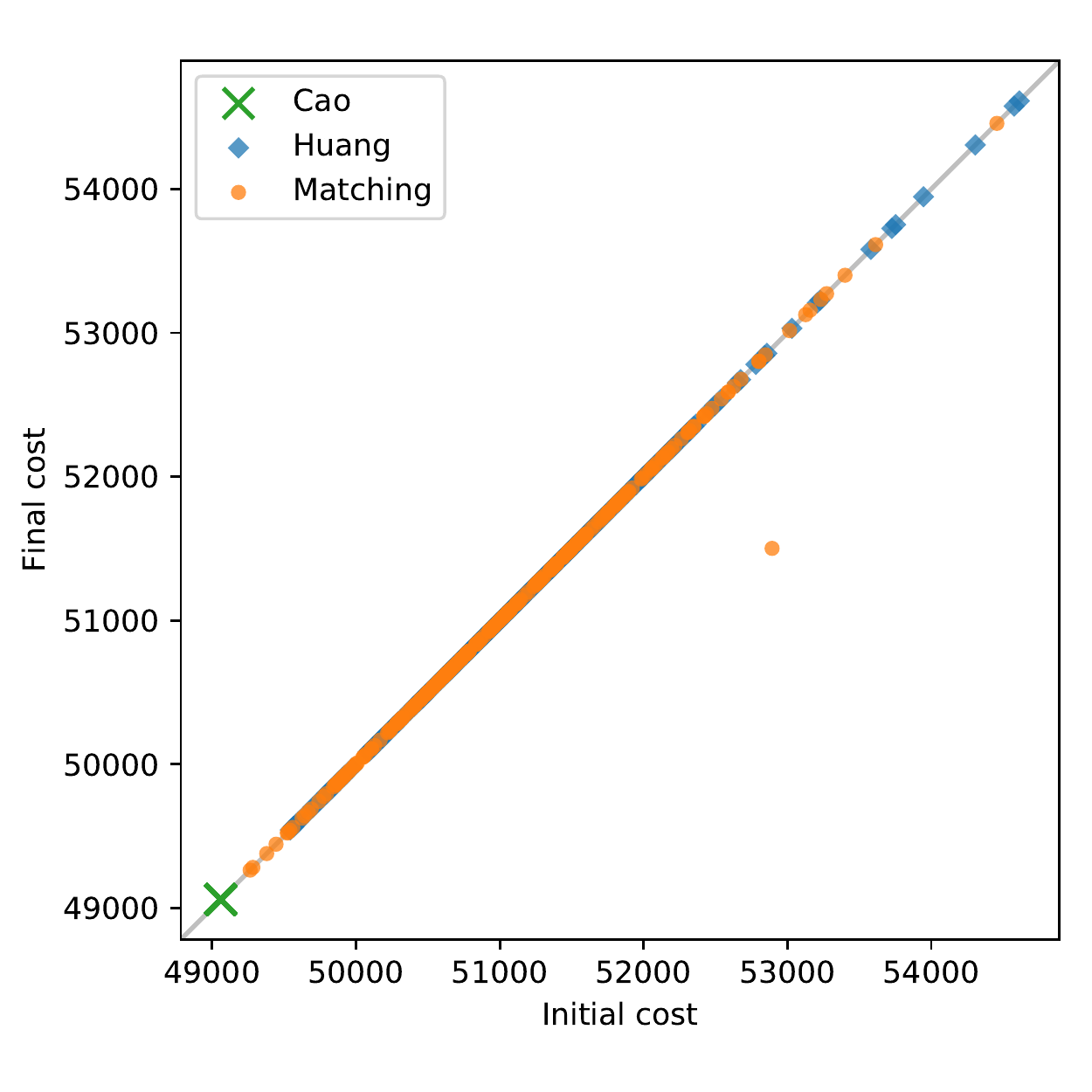}
        \caption{Scatter plot of initial and final costs.}
    \end{subfigure}
    \hfill%
    \begin{subfigure}{.5\textwidth}
        \includegraphics[width=\linewidth]{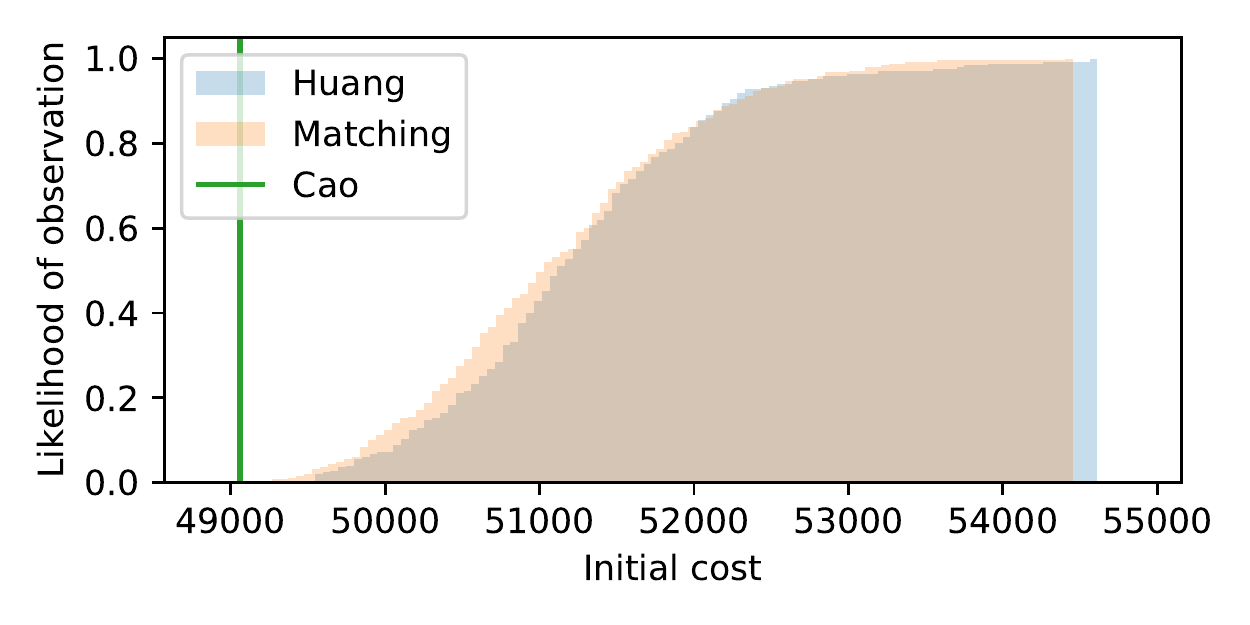}

        \includegraphics[width=\linewidth]{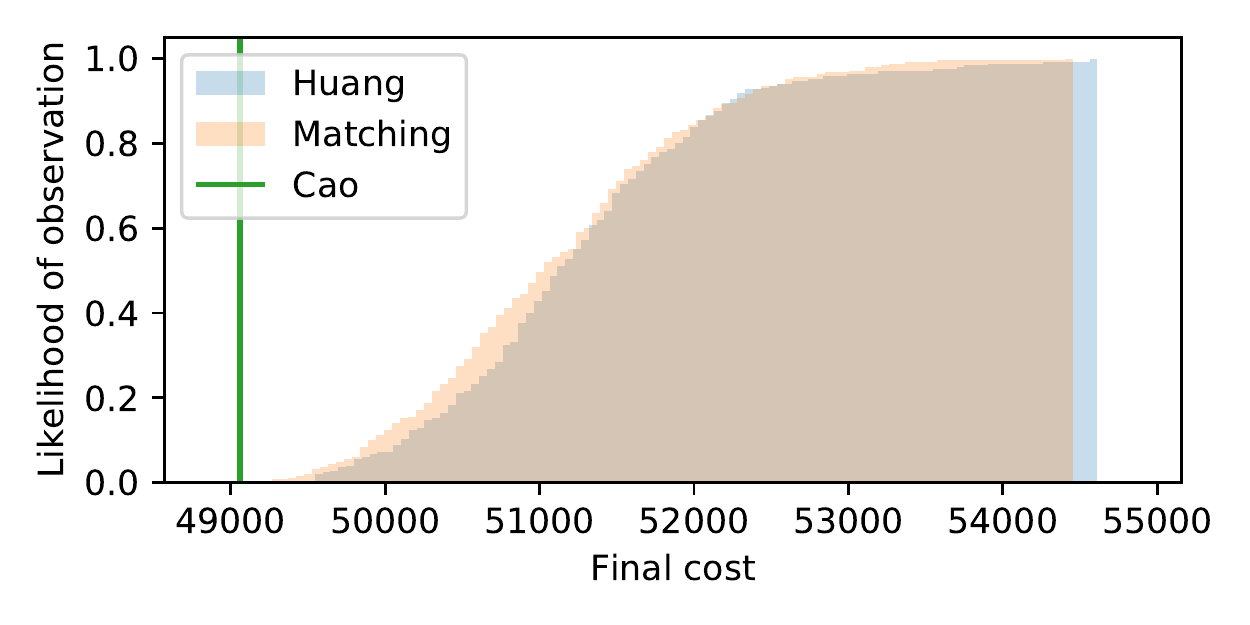}
        \caption{Empirical CDF plots for initial (top) and final (bottom)
                 costs.}
    \end{subfigure}
    \caption{Summative plots for the nursery dataset with \(k=5\).}%
    \label{fig:nursery_nclasses}
\end{figure}

\begin{figure}
    \begin{subfigure}{.5\textwidth}
        \includegraphics[width=\linewidth]{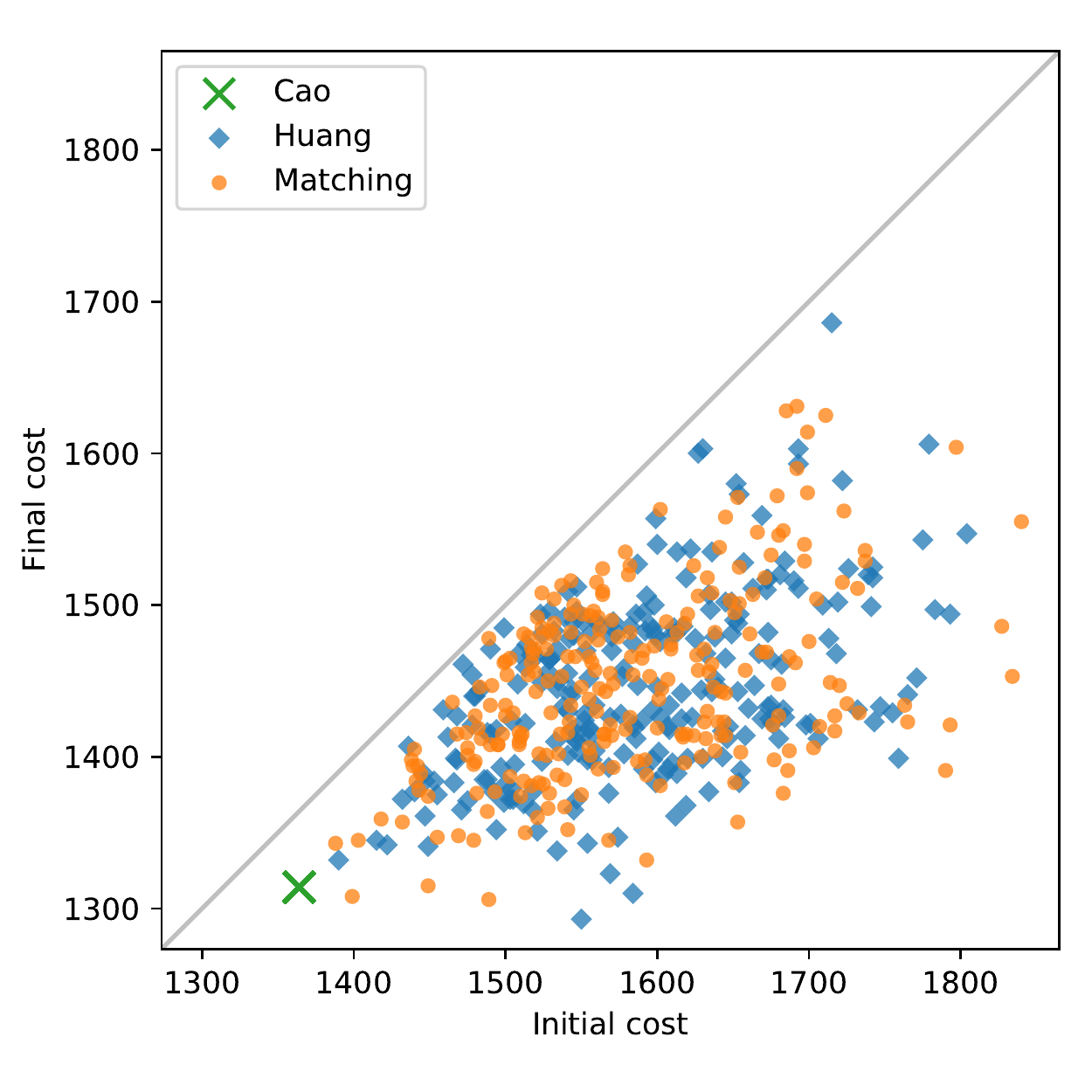}
        \caption{Scatter plot of initial and final costs.}
    \end{subfigure}
    \hfill%
    \begin{subfigure}{.5\textwidth}
        \includegraphics[width=\linewidth]{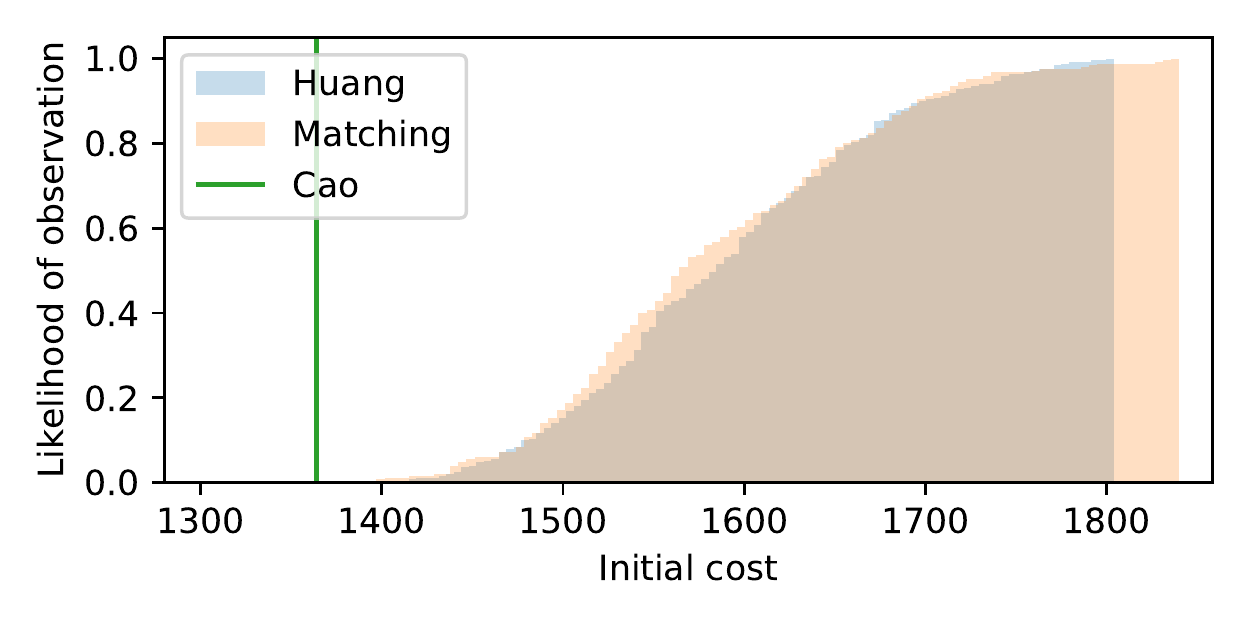}

        \includegraphics[width=\linewidth]{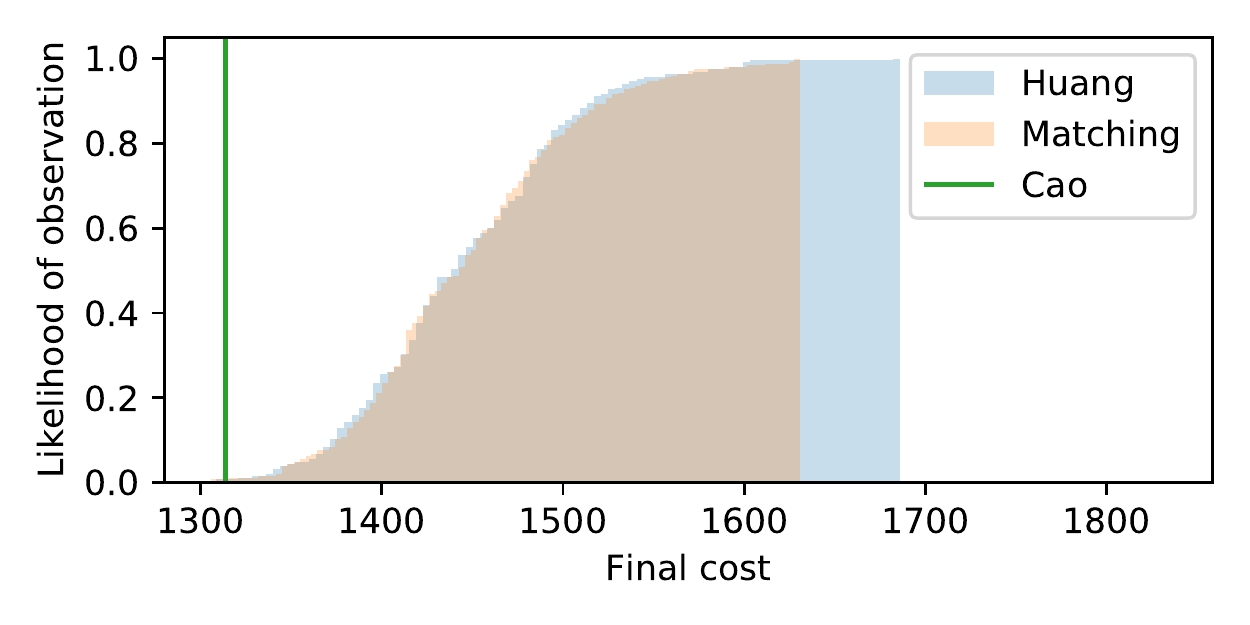}
        \caption{Empirical CDF plots for initial (top) and final (bottom)
                 costs.}
    \end{subfigure}
    \caption{Summative plots for the soybean dataset with \(k=15\).}%
    \label{fig:soybean_nclasses}
\end{figure}

Looking at these tables on their own, Cao's method is the superior
initialisation method on average: the means are substantially lower in terms of
initial and final cost; there is no deviation in these results; again, the total
computational time is a fraction of the other two methods. It is also apparent
that Huang's method and the proposed extension are very comparable on average.
As before, finer investigation will require finer visualisations.
Figures~\ref{fig:breast_cancer_nclasses}---\ref{fig:soybean_nclasses} show the
same plots as in the previous subsection except the number of clusters has been
taken to be the number of classes present in each dataset.

Figures~\ref{fig:breast_cancer_nclasses}~\&~\ref{fig:mushroom_nclasses} indicate
that a particular behaviour emerged during the runs of the \(k\)-modes
algorithm. Specifically, each solution falls into one of (predominantly) two
types: effectively no improvement on the initial clustering, or terminating at
some clustering with a cost that is bounded below across all such solutions.
Invariably, Cao's method achieves or approaches this lower bound and unless
Cao's method is used, these particular choices for \(k\) mean that the
performance of the \(k\)-modes algorithm is exceptionally sensitive to its
initial clustering. Moreover, the other two methods are effectively
indistinguishable in these cases and so if a robust solution is required, Cao's
method is the only viable option.

Figure~\ref{fig:nursery_nclasses} corresponds to the nursery dataset results
with \(k=5\). In this set of runs, the same pattern emerges as in
Figure~\ref{fig:nursery_knee} where sampling the initial centres from amongst
the most dense points (via Huang's method and the proposed) is an inferior
strategy to one considering the entire attribute space such as with Cao's
method. Again, no method is able to improve on the initial solution except for
one repetition with the matching initialisation method.

The primary conclusion from this analysis is that while Huang's method is
largely comparable to the proposed extension, there is no substantial evidence
from these use cases to use Huang's method over the one proposed in this work.
In fact, Figure~\ref{fig:soybean_nclasses} is the only instance where Huang's
method was able to outperform the proposed method. Other than this, the proposed
method consistently performing better (or as well as) Huang's method in terms of
minimal final costs and computational time over a number of runs in both the
cases where an external framework is imposed on the data (by choosing \(k\) to
be the number of classes) and not. Furthermore, though not discussed in this
work, the matching initialisation method has the scope to allow for expert or
prior knowledge to be included in an initial clustering by using some \textit{ad
hoc} preference list mechanism.

\subsection{Artificial datasets}\label{subsec:artificial}

Following on from the conclusions of the analysis thus far, the competition
between Cao's method and the proposed matching method may be studied more
deeply. All of the results leading up to this point were conducted using
benchmark datasets and while there are certainly benefits to comparing methods
in this way, it does not afford a rich understanding of how any of them perform
more generally. This stage of the analysis relies on a method for generating
artificial datasets introduced in~\cite{Wilde2019}. In essence, this method is
an evolutionary algorithm which acts on entire datasets to explore the space in
which potentially all possible datasets exist. The key component of this method
is an objective function that takes a dataset and returns a value that is to be
minimised; this function is referred to as the fitness function.

In order to reveal the nuances in the performance of Cao's method and the
proposed initialisation on a particular dataset, two cases are considered:
where Cao's method outperforms the proposed, and vice versa. Both cases use the
same fitness function --- with the latter using its negative --- which is
defined as follows:
\begin{equation}\label{eq:fitness}
    f\left(\mathcal X\right) = C_{\mathrm{cao}} - C_{\mathrm{match}}
\end{equation}
where \(C_{\mathrm{cao}}\) and \(C_{\mathrm{match}}\) are the final costs when a
dataset \(\mathcal X\) is clustered using Cao's method and the proposed matching
method respectively with \(k = 3\). For the sake of computational time, the
proposed initialisation was given 25 repetitions as opposed to the 250
repetitions in the remainder of this section. Apart from the sign of \(f\), the
dataset generation processes used identical parameters in each case and the
datasets considered here are all of comparable shape.

\begin{figure}
    \centering
    \includegraphics[width=\imgwidth]{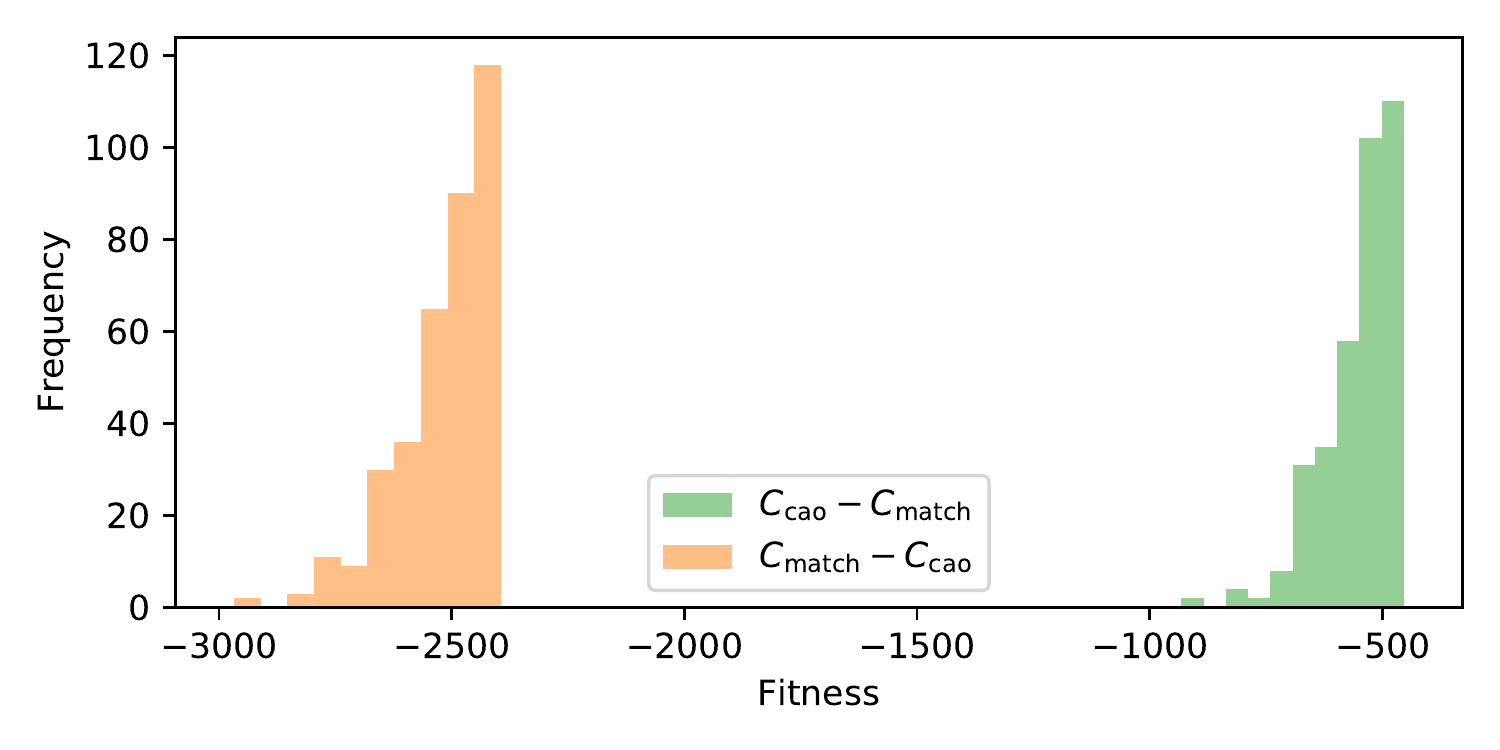}
    \caption{Histograms of fitness for the top performing percentile in each
             case.}\label{fig:fitness}
\end{figure}

This process yielded approximately 35,000 unique datasets for each case, and the
ensuing analysis only considers the top-performing percentile of datasets from
each. Figure~\ref{fig:fitness} shows the fitness distribution of the top
percentile in each case. It should be clear from~\eqref{eq:fitness} that large
negative values are preferable here. With that, and bearing in mind that the
generation of these datasets was parameterised in a consistent manner, it
appears that the attempt to outperform Cao's method proved somewhat easier. This
is indicated by the substantial difference in the locations of the fitness
distributions.

Given the quantity of data available, to understand the patterns that have
emerged, they must be summarised; in this case, univariate statistics are used.
Despite the datasets all being of similar shapes, there are some discrepancies.
With the number of rows this is less of an issue but any comparison of
statistics across datasets of different widths is difficult without prior
knowledge of the datasets. Moreover, there is no guarantee of contingency
amongst the attributes, and the comparison of more than a handful of variables
becomes complicated even when the attributes are identifiable. To combat this
and bring uniformity to the datasets, each dataset is represented as their first
principal component obtained via centred Principal Component Analysis
(PCA)~\cite{Jolliffe1986}. While some subtleties may be lost, this
representation captures the most important characteristics of each dataset in a
single variable meaning they can be compared directly.

Since the transformation by PCA is centred, all measures for central tendency
are moot. In fact, the mean and median are not interpretable here given that the
original data is categorical. As such, the univariate statistics used here
describe the spread and shape of the principal components, and are split into
two groups:
\begin{itemize}
    \item Central moments: variance, skewness and kurtosis.
    \item Empirical quantiles: interquartile range, lower decile and upper
        decile.
\end{itemize}

\begin{figure}
    \centering
    \begin{subfigure}{\imgwidth}
        \includegraphics[width=\linewidth]{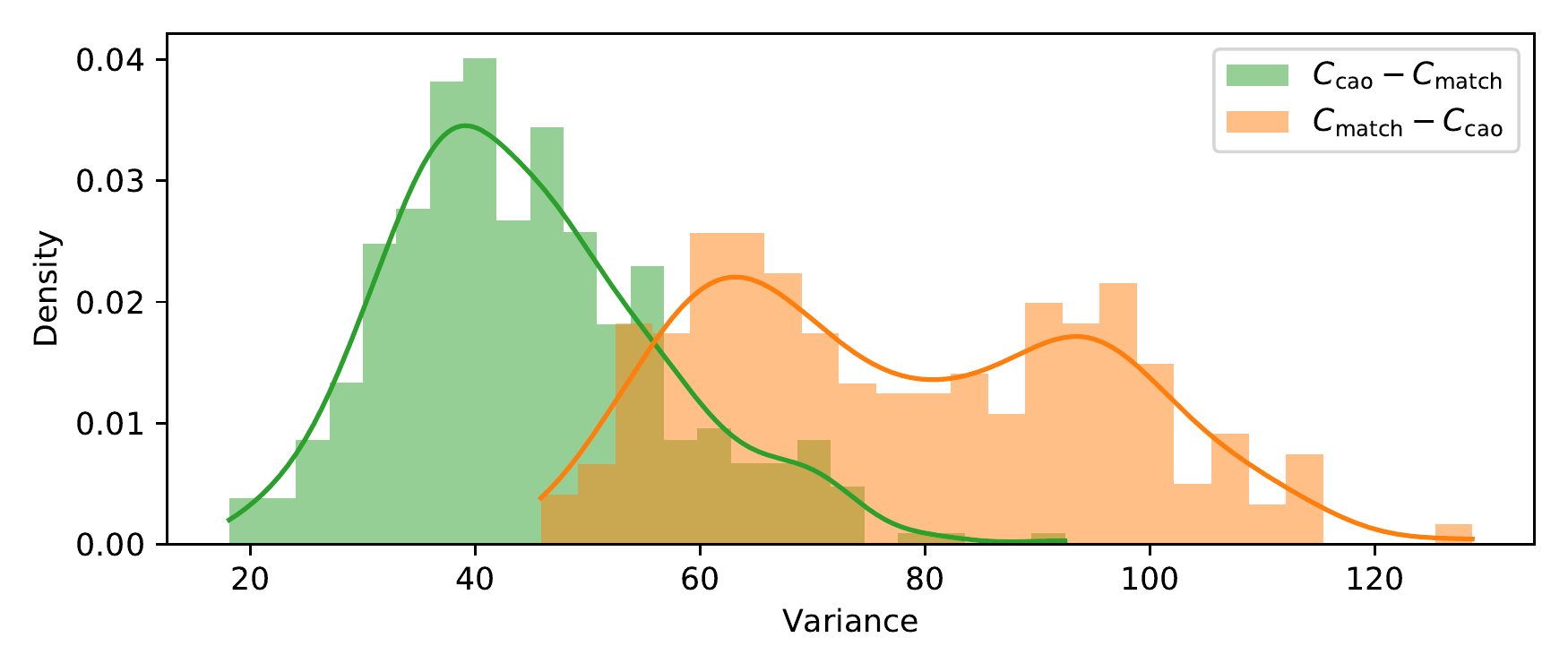}%
        \caption{}\label{fig:edo_variance}
    \end{subfigure}

    \begin{subfigure}{\imgwidth}
        \includegraphics[width=\linewidth]{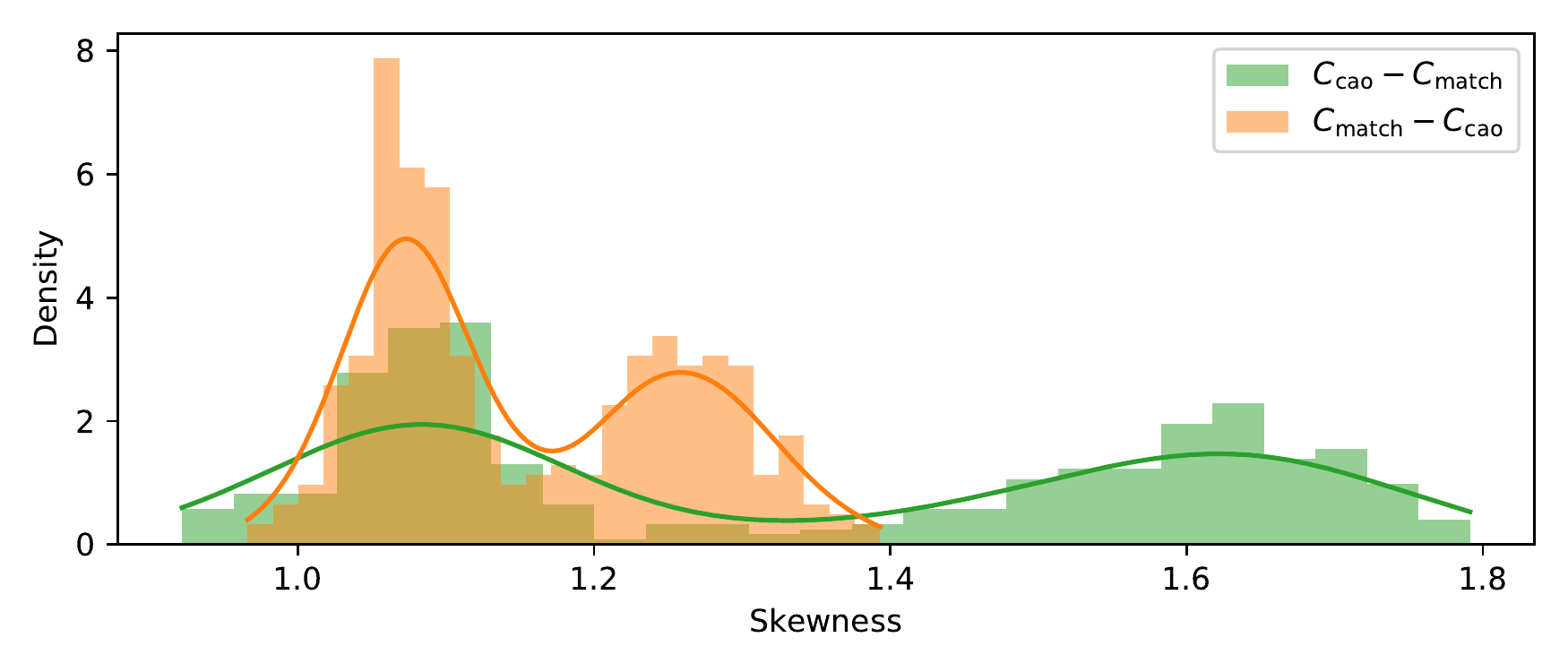}%
    \caption{}\label{fig:edo_skewness}
    \end{subfigure}
    
    \begin{subfigure}{\imgwidth}
        \includegraphics[width=\linewidth]{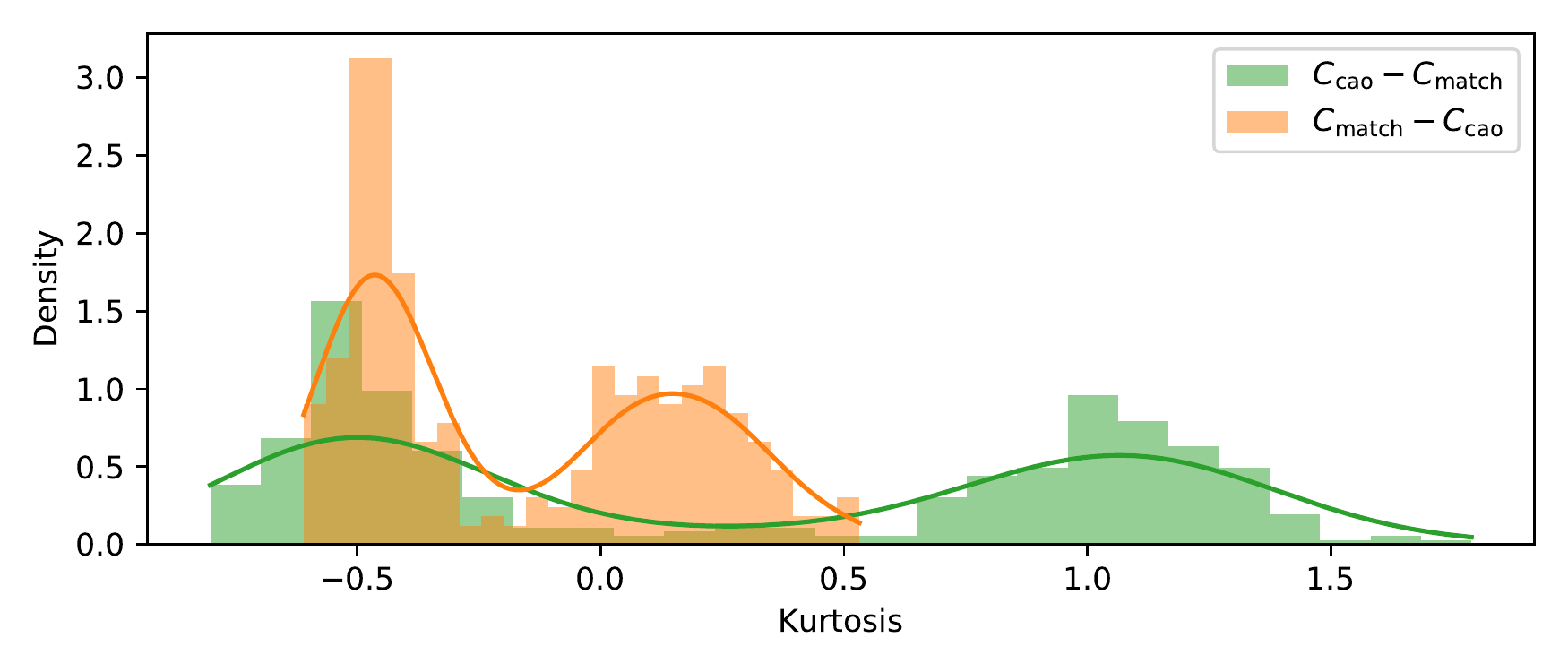}%
        \caption{}\label{fig:edo_kurtosis}
    \end{subfigure}
    \caption{Distribution plots for the (\subref{fig:edo_variance}) variance,
        (\subref{fig:edo_skewness}) skewness and (\subref{fig:edo_kurtosis})
        kurtosis of the first principal components in each
        case.}\label{fig:edo_moments}
\end{figure}

\begin{figure}
    \centering
    \begin{subfigure}{\imgwidth}
        \includegraphics[width=\linewidth]{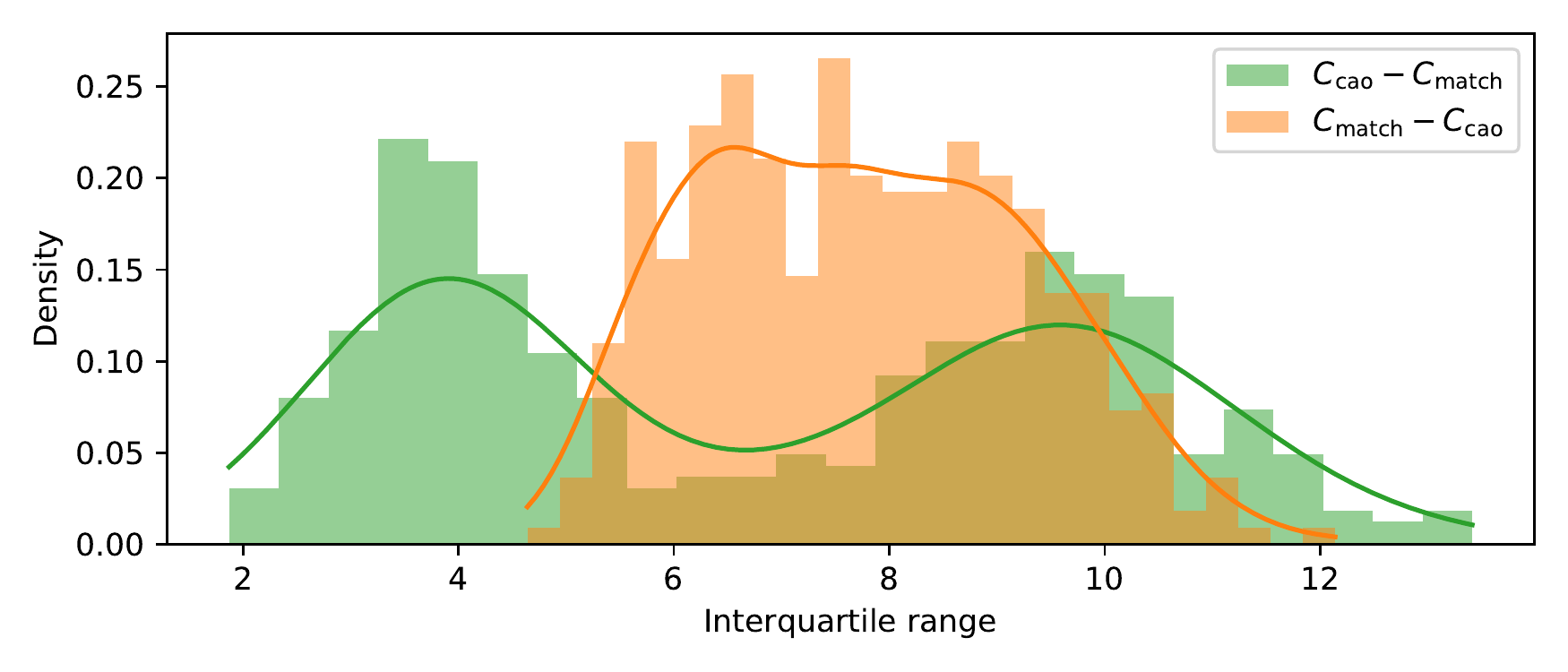}
        \caption{}\label{fig:edo_iqr}
    \end{subfigure}\vfill%

    \begin{subfigure}{\imgwidth}
        \includegraphics[width=\linewidth]{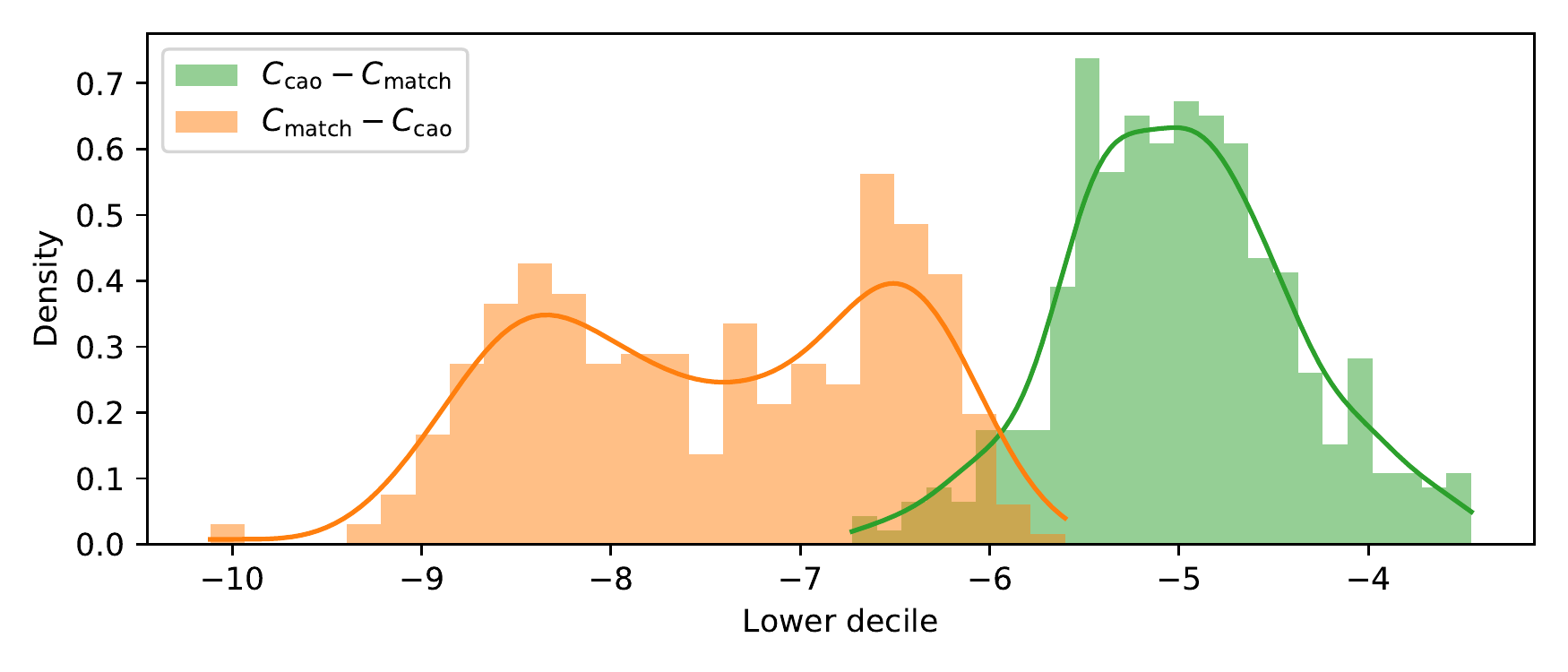}
        \caption{}\label{fig:edo_lower}
    \end{subfigure}\vfill%
    
    \begin{subfigure}{\imgwidth}
        \includegraphics[width=\linewidth]{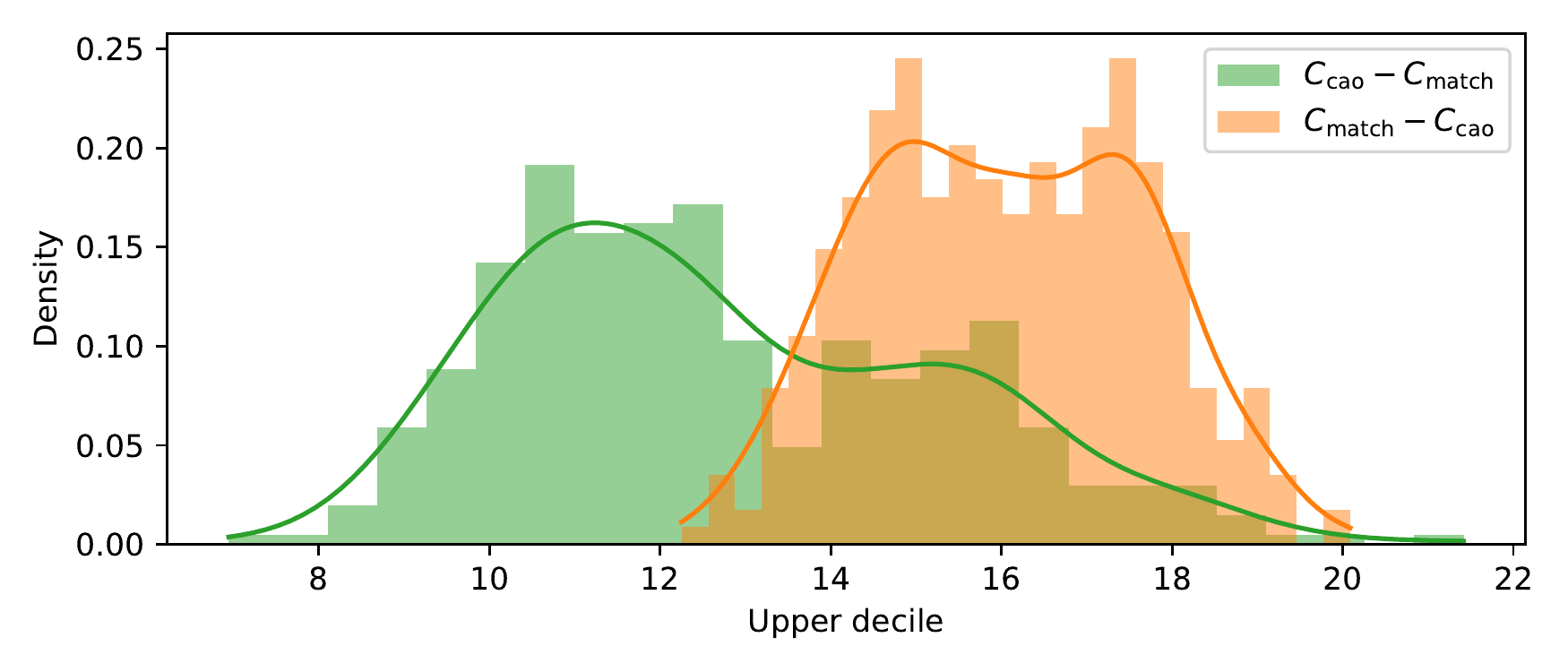}
        \caption{}\label{fig:edo_upper}
    \end{subfigure}
    \caption{Distribution plots for the (\subref{fig:edo_iqr}) interquartile
        range, (\subref{fig:edo_lower}) lower decile and
        (\subref{fig:edo_upper}) upper decile of the first principal components
        in each case.}\label{fig:edo_quantiles}
\end{figure}

Figures~\ref{fig:edo_moments}~\&~\ref{fig:edo_quantiles} show the distributions
of the six univariate statistics across all of the principal components in each
case. In addition to this, they show a fitted Gaussian kernel density
estimate~\cite{Bashtannyk2001} to accentuate the general shape of the
histograms. What becomes immediately clear from each of these plots is that for
datasets where Cao's method succeeds, the general spread of their first
principal component is much tighter than in the case where the proposed
initialisation method succeeds. This is particularly evident in
Figure~\ref{fig:edo_variance} where relatively low variance in the first case
indicates a higher level of density in the original categorical data.

The patterns in the quantiles further this. Although Figure~\ref{fig:edo_iqr}
suggests that the components of Cao-preferable datasets can have higher
interquartile ranges than in the second case, the lower and upper deciles tend
to be closer together as is seen in
Figures~\ref{fig:edo_lower}~\&~\ref{fig:edo_upper}. This suggests that despite
the body of the component being spread, its extremities are not.

In Figures~\ref{fig:edo_skewness}~\&~\ref{fig:edo_kurtosis}, the most notable
contrast between the two cases is the range in values for both skewness and
kurtosis. This supports the evidence thus far that individual datasets have
higher densities and lower variety (i.e.\ tighter extremities) when Cao's method
succeeds over the proposed initialisation. In particular, larger values of
skewness and kurtosis translate to high similarity between the instances in a
categorical dataset which is equivalent to having high density.

Overall, this analysis has revealed that if a dataset shows clear evidence of
high-density points, then Cao's method should be used over the proposed method.
However, if there is no such evidence, the proposed method is able to find a
substantially better clustering than Cao's method.

\section{Conclusion}\label{sec:conclusion}

In this paper a novel initialisation method for the \(k\)-modes was introduced
that built on the method set out in the seminal paper~\cite{Huang1998}. The new
method models the final `replacement' process in the original as an instance of
the Hospital-Resident Assignment Problem that may be solved to be mathematically
fair and stable.

Following a thorough description of the \(k\)-modes algorithm and the
established initialisation methods, a comparative analysis was conducted amongst
the three initialisations using both benchmark and artificial datasets. This
analysis revealed that the proposed initialisation was able to outperform both
of the other methods when the choice of \(k\) was optimised according to a
mathematically rigorous elbow method. However, the proposed method was unable to
beat Cao's method (established in~\cite{Cao2009}) when an external framework was
imposed on each dataset by choosing \(k\) to be the number of classes present.

The proposed method should be employed over Cao's when there are no hard
restrictions on what \(k\) may be, or if there is no immediate evidence that the
dataset at hand has some notion of high density. Otherwise, Cao's method remains
the most reliable initialisation in terms of computational time and final cost.

\bibliography{references}

\begin{thebibliography}{25}
\providecommand{\natexlab}[1]{#1}
\providecommand{\url}[1]{\texttt{#1}}
\expandafter\ifx\csname urlstyle\endcsname\relax
  \providecommand{\doi}[1]{doi: #1}\else
  \providecommand{\doi}{doi: \begingroup \urlstyle{rm}\Url}\fi

\bibitem[Arthur and Vassilvitskii(2007)]{Arthur2007}
D.~Arthur and S.~Vassilvitskii.
\newblock $k$-means++: {T}he advantages of careful seeding.
\newblock In \emph{Proceedings of the Eighteenth Annual ACM-SIAM Symposium on
  Discrete Algorithms}, SODA '07, pages 1027--1035, 2007.
\newblock ISBN 978-0-898716-24-5.
\newblock URL \url{http://dl.acm.org/citation.cfm?id=1283383.1283494}.

\bibitem[Bashtannyk and Hyndman(2001)]{Bashtannyk2001}
D.~M. Bashtannyk and R.~J. Hyndman.
\newblock Bandwidth selection for kernel conditional density estimation.
\newblock \emph{Computational Statistics and Data Analysis}, 36:\penalty0
  279--298, 2001.
\newblock ISSN 0167-9473.

\bibitem[Cao et~al.(2009)Cao, Liang, and Bai]{Cao2009}
F.~Cao, J.~Liang, and L.~Bai.
\newblock A new initialization method for categorical data clustering.
\newblock \emph{Expert Systems with Applications}, 36:\penalty0 10223--10228,
  2009.
\newblock URL \url{https://pdfs.semanticscholar.org/1955/
  c6801bca5e95a44e70ce14180f00fd3e55b8.pdf}.

\bibitem[Cao et~al.(2012)Cao, Liang, Li, Bai, and Dang]{Cao2012}
F.~Cao, J.~Liang, D.~Li, L.~Bai, and C.~Dang.
\newblock A dissimilarity measure for the $k$-modes clustering algorithm.
\newblock \emph{Knowledge-Based Systems}, 26:\penalty0 120--127, 2012.
\newblock \doi{10.1016/j.knosys.2011.07.011}.

\bibitem[Dua and Graff(2017)]{Dua2019}
D.~Dua and C.~Graff.
\newblock {UCI Machine Learning Repository}, 2017.
\newblock URL \url{http://archive.ics.uci.edu/ml}.

\bibitem[Erdil and Ergin(2017)]{Erdil2017}
A.~Erdil and H.~Ergin.
\newblock Two-sided matching with indifferences.
\newblock \emph{Journal of Economic Theory}, 171:\penalty0 268--292, 2017.
\newblock \doi{10.1016/j.jet.2017.07.002}.

\bibitem[Fuku et~al.(2006)Fuku, Namatame, and Kaizoji]{Fuku2006}
T.~Fuku, A.~Namatame, and T.~Kaizoji.
\newblock \emph{Collective Efficiency in Two-Sided Matching}, pages 115--126.
\newblock 2006.
\newblock \doi{10.1007/3-540-28547-4_10}.

\bibitem[Gale and Shapley(1962)]{Gale1962}
D.~Gale and L.~Shapley.
\newblock College admissions and the stability of marriage.
\newblock \emph{The American Mathematical Monthly}, 69\penalty0 (1):\penalty0
  9--15, 1962.
\newblock \doi{10.2307/2312726}.

\bibitem[Huang(1997{\natexlab{a}})]{Huang1997a}
Z.~Huang.
\newblock Clustering large data sets with mixed numeric and categorical values.
\newblock In \emph{The First {P}acific-{A}sia Conference on Knowledge Discovery
  and Data Mining}, pages 21--34, 1997{\natexlab{a}}.

\bibitem[Huang(1997{\natexlab{b}})]{Huang1997b}
Z.~Huang.
\newblock A fast clustering algorithm to cluster very large categorical data
  sets in data mining.
\newblock In \emph{Proceedings of the {SIGMOD} Workshop on Research Issues on
  Data Mining and Knowledge Discovery}, pages 1--8, 1997{\natexlab{b}}.

\bibitem[Huang(1998)]{Huang1998}
Z.~Huang.
\newblock Extensions to the $k$-means algorithm for clustering large data sets
  with categorical values.
\newblock \emph{Data Mining and Knowledge Discovery}, 2\penalty0 (3):\penalty0
  283--304, 1998.
\newblock \doi{10.1023/A:1009769707641}.

\bibitem[Iwama and Miyazaki(2016)]{Iwama2016}
K.~Iwama and S.~Miyazaki.
\newblock \emph{Stable Marriage with Ties and Incomplete Lists}, pages
  2071--2075.
\newblock Springer New York, 2016.
\newblock \doi{10.1007/978-1-4939-2864-4_805}.

\bibitem[Jiang et~al.(2016)Jiang, Liu, Du, and Sui]{Jiang2016}
F.~Jiang, G.~Liu, J.~Du, and Y.~Sui.
\newblock Initialization of $k$-modes clustering using outlier detection
  techniques.
\newblock \emph{Information Sciences}, 332:\penalty0 167--183, 2016.
\newblock \doi{10.1016/j.ins.2015.11.005}.

\bibitem[Jolliffe(1986)]{Jolliffe1986}
I.~T. Jolliffe.
\newblock \emph{Principal Component Analysis and Factor Analysis}, pages
  115--128.
\newblock Springer New York, 1986.
\newblock \doi{10.1007/978-1-4757-1904-8_7}.

\bibitem[Kwanashie et~al.(2015)Kwanashie, Irving, Manlove, and
  Sng]{Kwanashie2015}
A.~Kwanashie, R.~W. Irving, D.~F. Manlove, and C.~T.~S. Sng.
\newblock Profile-based optimal matchings in the student/project allocation
  problem.
\newblock In \emph{Combinatorial Algorithms}, pages 213--225, 2015.
\newblock \doi{10.1007/978-3-319-19315-1_19}.

\bibitem[Manlove et~al.(2002)Manlove, Irving, Iwama, Miyazaki, and
  Morita]{Manlove2002}
D.~F. Manlove, R.~W. Irving, K.~Iwama, S.~Miyazaki, and Y.~Morita.
\newblock Hard variants of stable marriage.
\newblock \emph{Theoretical Computer Science}, 276\penalty0 (1):\penalty0
  261--279, 2002.
\newblock \doi{10.1016/S0304-3975(01)00206-7}.

\bibitem[M{\'e}moli(2011)]{Memoli2011}
F.~M{\'e}moli.
\newblock Metric structures on datasets: Stability and classification of
  algorithms.
\newblock In \emph{Computer Analysis of Images and Patterns}, pages 1--33.
  Springer Berlin Heidelberg, 2011.
\newblock \doi{10.1007/978-3-642-23678-5_1}.

\bibitem[Ng et~al.(2007)Ng, Li, Huang, and He]{Ng2007}
M.~K. Ng, M.~J. Li, J.~Z. Huang, and Z.~He.
\newblock On the impact of dissimilarity measure in $k$-modes clustering
  algorithm.
\newblock \emph{IEEE Transactions on Pattern Analysis and Machine
  Intelligence}, 29\penalty0 (3):\penalty0 503--507, 2007.
\newblock \doi{10.1109/TPAMI.2007.53}.

\bibitem[Olaode et~al.(2014)Olaode, Naghdy, and Todd]{Olaode2014}
A.~Olaode, G.~Naghdy, and C.~Todd.
\newblock Unsupervised image classification by {P}robabilistic {L}atent
  {S}emantic {A}nalysis for the annotation of images.
\newblock In \emph{International Conference on Digital Image Computing:
  Techniques and Applications}, 2014.
\newblock \doi{10.13140/2.1.1909.4086}.

\bibitem[Roth(1984)]{Roth1984}
A.~Roth.
\newblock The evolution of the labor market for medical interns and residents:
  A case study in game theory.
\newblock \emph{Journal of Political Economy}, 92\penalty0 (6):\penalty0
  991--1016, 1984.
\newblock \doi{10.1086/261272}.

\bibitem[Satopaa et~al.(2011)Satopaa, Albrecht, Irwin, and
  Raghavan]{Satopaa2011}
V.~Satopaa, J.~Albrecht, D.~Irwin, and B.~Raghavan.
\newblock Finding a `kneedle' in a haystack: Detecting knee points in system
  behavior.
\newblock In \emph{Proceedings of the 2011 31st International Conference on
  Distributed Computing Systems Workshops}, pages 166--171, 07 2011.
\newblock \doi{10.1109/ICDCSW.2011.20}.

\bibitem[Schaeffer(2007)]{Schaeffer2007}
S.~E. Schaeffer.
\newblock Graph clustering.
\newblock \emph{Computer Science Review}, 1\penalty0 (1):\penalty0 27--64,
  2007.
\newblock ISSN 1574-0137.
\newblock \doi{10.1016/j.cosrev.2007.05.001}.

\bibitem[Sharma and Gaud(2015)]{Sharma2015}
N.~Sharma and N.~Gaud.
\newblock $k$-modes clustering algorithm for categorical data.
\newblock \emph{International Journal of Computer Applications}, 127\penalty0
  (17):\penalty0 1--6, 2015.
\newblock \doi{10.5120/ijca2015906708}.

\bibitem[{The Matching library developers}(2019)]{Matching1.1}
{The Matching library developers}.
\newblock Matching: v1.1, 2019.
\newblock URL \url{http://dx.doi.org/10.5281/zenodo.2711847}.

\bibitem[Wilde et~al.(2019)Wilde, Knight, and Gillard]{Wilde2019}
H.~Wilde, V.~Knight, and J.~Gillard.
\newblock Evolutionary dataset optimisation: learning algorithm quality through
  evolution.
\newblock \emph{Applied Intelligence}, 2019.
\newblock \doi{10.1007/s10489-019-01592-4}.

\end{thebibliography}

\end{document}